\documentclass[11pt]{article}
\usepackage[margin=1in]{geometry}

\usepackage[T1]{fontenc}
\usepackage{lmodern}
\usepackage{amsmath,amssymb}
\usepackage{graphicx}
\usepackage{booktabs}
\usepackage{url}
\usepackage[hidelinks]{hyperref}
\usepackage{float}

\usepackage[natbibapa]{apacite}   %
\makeatletter\NAT@longnamesfalse\makeatother   %

\newcommand{\runin}[1]{\smallskip\noindent\textit{#1.}\enspace}

\newif\ifrepo
\repofalse
\newcommand{\repourl}{https://github.com/ShuyaoGao/sudoku-skill-ladder-snn}

\begin{document}

\title{What must be built in? Small spiking networks learning $4\times4$ Sudoku across levels of skill encapsulation}
\author{Shuyao Gao\thanks{aSSIST University, Seoul, South Korea.
Email: \texttt{gaoshuyao.mk@gmail.com}.\ifrepo{} Code and data: \url{\repourl}.\fi}
\and Minghao Huang\thanks{aSSIST University, Seoul, South Korea.
Corresponding author. Email: \texttt{mhuang@assist.ac.kr}.}}
\date{}
\maketitle

\begin{abstract}
Hand-wired spiking networks can solve Sudoku, but their constraints are
designed, not learned. We ask how much of a solver must be built in before
a small spiking network can learn the rest. In $4\times4$ Sudoku every
uniquely solvable puzzle yields to waves of single-candidate inference, which gives an exact
difficulty scale. With learning rates tuned for every network on held-out
validation grids, and tested on solution grids never seen in training, a
learned one-wave operator inside a provided loop solves 98.7\% of puzzles.
Keeping the loop but training on full solutions, or giving internal
recurrence the same per-wave targets, each reaches 69\%; networks that
output full solutions largely recall training grids and fail mainly on the
grid type that admits deep puzzles. Non-spiking counterparts show the same
ordering, although with clean input per-wave targets and internal
recurrence come within 6 points of the provided loop. Under feedback
alone, a spiking policy that can represent a zero-lookahead rule learns it
to 61\% from outcome reward and to 90\% from rule-aligned reward. The
spiking operator matches its non-spiking counterparts on clean input. It also still solves 93\% of puzzles when
10\% of its input bits flip at random over time, but so does a non-spiking
network that, like a spiking neuron, integrates its input before the
nonlinearity: the robustness comes from that integration, not from spikes.
\end{abstract}


\section{Introduction}
\label{sec:introduction}
Hand-wired networks of spiking neurons can solve Sudoku.
\citet{habenschuss2013} wired overlapping winner-take-all circuits so that
a network of stochastically firing neurons samples configurations of a
$9\times9$ puzzle and spends most of its time near solutions, and
\citet{jonke2016} turned the construction into a design method for
constraint satisfaction problems. Related circuits run on neuromorphic
hardware \citep{mostafa2015,fonsecaguerra2017}. In all of these the
constraints are designed into the connectivity. A different line of work
shows that cultured neurons embedded in a closed stimulation loop change
their activity in a way that improves play in a simple arcade game
\citep{kagan2022}, a result whose interpretation remains debated
\citep{balci2023}. Between a solver that is built and a network that is
taught only by consequences lies a question that neither line answers: how
much of a multi-step solver has to be built in before a small spiking
network can learn the remainder? We call the part of a solver that is
provided by the experimenter, rather than learned, its degree of
\emph{skill encapsulation}.

Existing approaches leave this question open for different reasons.
First, hand-wired spiking solvers show what spiking dynamics can
represent, and they do so convincingly, but they do not learn. Second,
learned neural solvers for Sudoku \citep{palm2018,wang2019,yang2023} and
recent recursive reasoning models \citep{wang2025hrm,jolicoeur2025} are
non-spiking and grade difficulty by the number of givens rather than by
the length of the inference chain. Third, work on algorithmic reasoning
shows that supervising intermediate steps makes such tasks learnable
\citep{xu2020,velickovic2022,wies2023,shah2024}, and recurrent networks
can extrapolate from easy to hard instances
\citep{schwarzschild2021,bansal2022}, but again without spikes and without
a controlled decomposition of the solver. Fourth, reward-modulated
learning in spiking networks
\citep{izhikevich2007,fremaux2016,bellec2020,tang2021popsan} is usually
studied on tasks in which a single decision is followed closely by its
outcome, so it is rarely tested what limits learning when consequences
are delayed.

This paper measures how learning depends on how much of the solver is
provided. We use $4\times4$ Sudoku because its puzzle space can be
enumerated and every puzzle has an exact depth: the number of waves of
single-candidate inference needed to finish it. We decompose the solver
into four components and build a ladder of six levels that differ in
which components are provided. We train spiking networks and matched
non-spiking networks with back-propagation through time (BPTT) under
supervision, and spiking policies with REINFORCE under environmental
feedback. Because a $4\times4$ puzzle has only 288 possible solutions, we
test on puzzles whose solution grids were held out of training, and we
check the result against the symmetries of the task. A zero-lookahead
rule that always fills the cell with the fewest legal digits solves every
puzzle, so the step-by-step task never requires lookahead. The main study
was pre-specified in a version-controlled record; it replaced an
exploratory study whose conclusions it partly overturned, and further
controls were added after the pre-specified analysis. The main study used
one learning rate for all networks; because that rate proved too low for
most of them, every supervised arm was re-run with its own learning rate,
chosen by a rule recorded before the runs. We report the tuned results and
list the untuned ones in the Appendix.

The main contributions of this study are:
\begin{enumerate}
    \item \textbf{A depth-calibrated testbed and an encapsulation ladder,
    evaluated on held-out solution grids.} Six levels (L0 to L5) are
    compared, with five seeds, non-spiking counterparts, a test set drawn
    and hashed before training, and checks against the symmetries of the
    grids, which fall into only two essentially different types.
    \item \textbf{What must be built in.} A learned one-wave operator
    generalizes almost completely when the loop is provided. The loop
    without one-wave targets, or one-wave targets without the loop, each
    help, but for spiking networks each falls about 30 points short and
    leaves a large gap between seen and held-out grids. The known benefit of
    intermediate supervision is largest together with an external loop,
    and the shortfall of the other arms lies mainly on the grid type that
    admits deep puzzles.
    \item \textbf{What limits learning from feedback, and what spikes
    contribute.} A spiking policy that can represent the zero-lookahead rule
    learns it only partly from outcome reward and much better from a reward
    aligned with the rule; stepwise interaction matters more than reward
    granularity. With learning rates tuned, the spiking L3 and L5 were as
    accurate as their non-spiking counterparts on clean input, while
    L3$^{*}$ and the recurrent networks were less accurate. Under
    time-varying input noise the spiking operator was far more robust than
    non-spiking networks that apply the nonlinearity before averaging over
    time, but no more robust than one that integrates its input first.
\end{enumerate}

\section{Related work}
\label{sec:related}

\subsection{Spiking networks for constraint satisfaction}
Stochastic spiking networks can be read as samplers from a probability
distribution over network states \citep{buesing2011}.
\citet{habenschuss2013} used this view to build circuits whose stationary
distribution concentrates on solutions of Sudoku and other constraint
problems, and \citet{jonke2016} generalized the construction. Event-based
and neuromorphic implementations \citep{mostafa2015,fonsecaguerra2017}
showed that such circuits run efficiently in hardware. The present paper
differs in two respects: (i) the constraint-handling circuitry is
learned, to a degree that we vary systematically; (ii) difficulty is
measured by the exact length of the inference chain rather than by puzzle
size.

\subsection{Learned neural solvers and algorithmic reasoning}
Recurrent relational networks reach high accuracy on $9\times9$ Sudoku by
passing messages between cells for a fixed number of steps, with a loss
at every step \citep{palm2018}. Differentiable optimization layers
\citep{amos2017} and SATNet \citep{wang2019} embed a solver in a network;
SATNet can learn the rules of Sudoku from examples, although its ability
to do so from raw visual input has been questioned \citep{chang2020}.
Recurrent transformers learn constraint satisfaction problems including
Sudoku \citep{yang2023}, and recent recursive models reach high accuracy
on hard Sudoku with small networks applied many times
\citep{wang2025hrm,jolicoeur2025}. A network learns an algorithm more
easily when its structure aligns with the algorithm's steps
\citep{xu2020}; supervising the intermediate states of an algorithm
(hints) helps neural executors \citep{velickovic2022}; decomposing a task
into supervised sub-tasks can make an unlearnable task learnable
\citep{wies2023}; and transformers trained on solver-ordered steps solve
logic puzzles including Sudoku \citep{shah2024}. Adaptive computation
time \citep{graves2016} lets a recurrent network choose how many steps to
spend on an input, and recurrent networks trained on small instances can
be run longer to solve larger ones \citep{schwarzschild2021,bansal2022}.
Our comparison of one-wave and full-solution targets is an instance of
intermediate supervision. What we add is a separation of that supervision
from the loop that applies it, in spiking networks, on an exact depth
scale and with generalization measured on held-out solution grids.

\subsection{Training spiking networks}
Surrogate gradients make BPTT usable in networks of spiking neurons
\citep{neftci2019,zenke2018}, and the approach is robust to the choice of
surrogate function \citep{zenke2021}. Standard benchmarks for this family
of methods are sensory classification tasks \citep{cramer2022}. For
learning from reward, three-factor rules combine local eligibility traces
with a global modulatory signal \citep{izhikevich2007,fremaux2016},
e-prop approximates BPTT with local traces \citep{bellec2020}, and
population-coded spiking actors have been trained with deep reinforcement
learning for continuous control \citep{tang2021popsan}. We use
surrogate-gradient BPTT together with REINFORCE \citep{williams1992}.
These are not local rules, so our results do not predict what a local
rule would learn.

\subsection{Closed-loop learning in cultured neurons}
\citet{kagan2022} reported that cortical cultures receiving predictable
stimulation after a hit and unpredictable stimulation after a miss
improved at a Pong-like game. The feedback in that experiment is
structured stimulation, not a scalar reward, and the interpretation of
the result has been criticized \citep{balci2023}. We borrow only the
experimental motif of a network embedded in a loop with an environment
that returns consequences of its actions. Our feedback-trained policies
receive a scalar reward and are trained with gradient methods; they are
not a model of the culture experiment.

\subsection{Summary of the gap}
Work on spiking constraint solvers is typically representational: the
circuit is designed and its dynamics are analyzed. Work on learned neural
solvers and algorithmic reasoning is typically non-spiking, grades
puzzles by the number of givens, and rarely separates new puzzles from
new solution structures. Work on reward learning in spiking networks has
examined credit assignment over time but less often on tasks where the
delay between a wrong decision and its consequence can be measured and
manipulated. The contribution of the present paper is to connect these
strands on one task with an exact difficulty scale: we decompose a Sudoku
solver into components, vary which components are provided, and measure
on held-out solution grids where learning succeeds and where it fails
under supervision and under feedback.

\section{Methods}
\label{sec:methodology}

\subsection{Task and depth scale}
\begin{figure}[t]
    \centering
    \includegraphics[width=\textwidth]{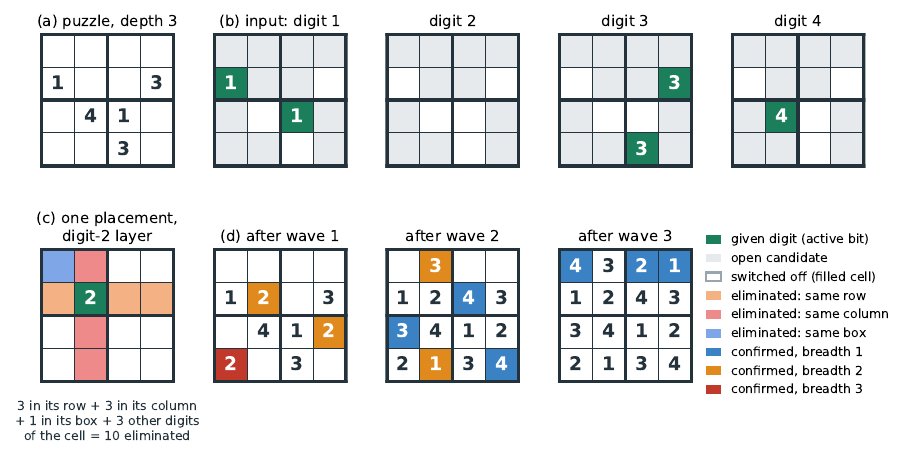}
    \caption{The task and its depth scale, shown on a depth-3 puzzle.
    (a) The puzzle with five givens. (b) The network input: 64 candidate
    bits drawn as four digit layers. Green bits belong to given digits; a
    filled cell switches off its other three digits (white); every empty
    cell keeps all four candidates open (gray), so no elimination is
    provided. (c) What a single placement implies, drawn for digit 2 at
    row 2, column 2 of an empty board: it eliminates the same digit in
    three cells of its row (orange), three of its column (red) and the one
    remaining cell of its box (blue), plus the other three digits of its
    own cell, 10 candidates in all. (d) The board after each wave of
    single-candidate inference. Colored cells are confirmed in that wave
    and are colored by integration breadth: blue when one unit of the cell
    already holds three digits (breadth 1), orange when two units must be
    combined (breadth 2), red when all three units are needed (breadth 3).
    Three waves complete the puzzle, so its depth is 3.}
    \label{fig:task}
\end{figure}

There are 288 complete $4\times4$ grids. A brute-force uniqueness count
over all $288\times2^{16}$ grid and mask pairs gives 13,579,680 uniquely
solvable puzzles (13,579,392 excluding the 288 complete grids), and an
exhaustive pass confirms that iterated naked singles solve every one of
them. We define a wave as one parallel application of naked-single
inference to the current board: every empty cell with exactly one
remaining candidate is confirmed at once. The depth of a puzzle is the
number of waves needed to complete it (one to five). We draw a stratified
pool of 54,000 puzzles (12,000, 20,000, 12,000, 6,000 and 4,000 at depths
1 to 5); the generator chooses the number of givens adaptively to fill
these quotas, so the pool is not a uniform sample of the puzzle space.
Integration breadth is the number of units (row, column, box) that must
be combined to certify a cell: breadth 1 when one unit of the cell already
holds three digits, breadth 2 when the digits of two units together cover
three values, and breadth 3 otherwise. Fig.~\ref{fig:task} illustrates
these definitions on a depth-3 test puzzle.

\subsection{Splits by solution grid and the symmetries of the grids}
Because there are only 288 solution grids, a split by puzzle places every
test grid in the training set as well, and a network that maps givens to
a full solution could identify the grid instead of inferring it. The main
split therefore holds out grids. With a fixed seed we held out 58 of the
288 grids; puzzles with these grids form the test pool, and no puzzle with
a held-out grid is used for training or validation. The main test set
contains 300 puzzles per depth from held-out grids. A second test set of
the same size contains new puzzles whose grids occur in training;
comparing the two measures how much a network relies on material seen in
training. Both test sets exclude every puzzle used for evaluation in the
exploratory study. Validation uses 3,000 puzzles and training 39,081
puzzles, all from training grids. The split was drawn and its hash
recorded before any training in the main study. The two test sets are
matched in difficulty: the mean number of empty cells per depth differs
by at most 0.3 between them.

The grids have many symmetries. The group generated by relabeling the
four digits (24 elements) and by the geometric symmetries (swapping rows
within a band, swapping the two bands, the same for columns, and
transposition; 128 elements) has 3,072 elements. Relabeling alone groups
the 288 grids into 12 classes of 24; the full group leaves only two
essentially different grid types, with 96 and 192 grids. By the
orbit--stabilizer theorem, each 96-type grid is mapped onto itself by 32
of the 3,072 symmetries and each 192-type grid by 16, which we confirmed
by enumeration: the smaller type is twice as self-symmetric. We also
verified by exhaustive enumeration that the two types differ in depth: no
puzzle built on a 96-type grid needs more than three waves, whereas
192-type grids admit puzzles of depth 4 and 5 (for one representative
grid of each type, the uniquely solvable puzzles number 41,401 and 50,027,
of which 0 and 2,720 have depth 4 or 5; $96\times41{,}401 +
192\times50{,}027 = 13{,}579{,}680$). A random holdout need not keep a
relabeling of every held-out grid in training, but we checked that ours
does: all 12 relabeling classes of the held-out grids also occur among
the training grids, so every held-out grid is a relabeling of some
training grid, and even a held-out relabeling class is a geometric
transform of training classes. Held-out
grids protect against recalling exact grids by networks without spatial
weight sharing; they do not test transfer to new solution structures in a
stronger sense. We report two further splits as checks: a class split (3
of the 12 relabeling classes for testing, 1 for validation, 8 for
training; 33,812 training puzzles) and a type split (train on 192-type
grids, test on 96-type grids, which limits testing to depths 1 to 3).

Table~\ref{tab:task} summarizes the main test set. Deeper puzzles have
more empty cells, which we control for in Section~\ref{sec:ladder-res}. In a
depth-1 puzzle every legal first move is correct. From depth 2 on, a
growing share of legal moves is wrong, and a wrong move is exposed only
when some later board contains an empty cell with no legal digit. We
measured this delay by continuing the game after each wrong legal first
move with the fewest-legal-digits rule, in the environment of
Section~\ref{sec:feedback-env}, and counting the moves until the dead end.
Two non-learning policies bracket the task. The fewest-legal-digits rule
solves every puzzle; this follows from the naked-singles property,
because on any board reached by correct moves some empty cell has exactly
one legal digit, and the rule fills such a cell. A random policy that
picks uniformly among all legal (cell, digit) pairs has the exact expected
solve rate shown in the table, computed by recursion over each puzzle's
move tree.

\begin{table}[t]
\caption{Structure of the held-out-grid test set and two non-learning reference policies.}
\label{tab:task}
\centering
\begin{tabular}{lccccc}
\toprule
Depth (waves) & 1 & 2 & 3 & 4 & 5 \\
\midrule
Test puzzles & 300 & 300 & 300 & 300 & 300 \\
Mean empty cells & 5.5 & 7.7 & 10.0 & 10.6 & 11.1 \\
Wrong share of legal first moves (\%) & 0.0 & 25.9 & 44.0 & 49.7 & 52.2 \\
Wrong first moves with a delayed dead end (\%) & -- & 0.0 & 27.7 & 52.2 & 60.9 \\
Mean delay to dead end (moves) & -- & 0.00 & 0.48 & 1.24 & 1.64 \\
Fewest-legal-digits rule, solved (\%) & 100 & 100 & 100 & 100 & 100 \\
Random legal move, expected solved (\%) & 100 & 49.9 & 16.7 & 10.2 & 7.0 \\
\bottomrule
\end{tabular}
\end{table}

\runin{Why $4\times4$ puzzles rather than $9\times9$}
The standard puzzle size in work on spiking constraint solvers is
$9\times9$. We chose $4\times4$ for three reasons. First, the space can be
enumerated, so depth is an exact property of every puzzle rather than an
estimate from a solver trace. Second, every puzzle is solvable by
single-candidate inference, which lets the encapsulation ladder be defined
over one fixed set of components. Third, networks of a few hundred
neurons can be trained on a laptop, which makes multi-seed comparisons
across many arms affordable. The cost is that the task never requires
inference beyond single candidates, and that its small number of solution
grids makes recall a real alternative to inference. We view $9\times9$
puzzles that require pairs or case analysis as a complementary, not
competing, direction.

\subsection{Encapsulation ladder}
\begin{figure}[t]
    \centering
    \includegraphics[width=\textwidth]{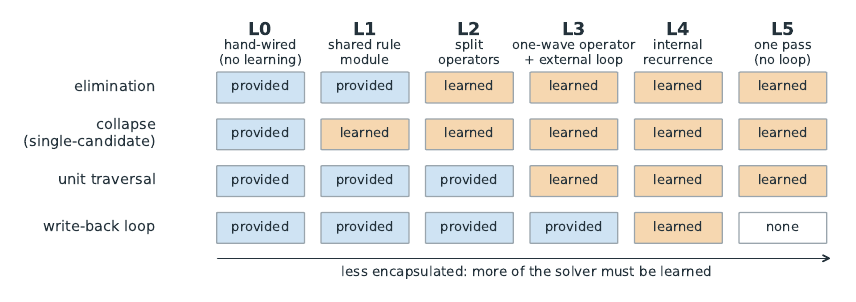}
    \caption{The encapsulation ladder. Columns are levels L0 to L5; rows are
    the four components of single-candidate solving. Blue cells are
    components provided by the experimenter (hand-built or run by the
    harness); orange cells are components the network must learn; the white
    cell marks that L5 has no loop of any kind. In L4 the loop is internal
    recurrence, learned without intermediate targets (L4-ds adds them).}
    \label{fig:ladder}
\end{figure}

Solving a puzzle by single-candidate inference has four components:
elimination (a placed digit rules out conflicting candidates), collapse (a
cell with one surviving candidate is confirmed), traversal of the units
that tells each candidate which cells are its peers, and a write-back
loop that feeds confirmed digits into the next wave. The ladder in
Fig.~\ref{fig:ladder} varies which of these are provided and which are
learned.
\begin{itemize}
    \item \textbf{L0} is a hand-wired spiking solver (Section~\ref{sec:l0}).
    \item \textbf{L1} receives the exact candidate set of each empty cell
    from the harness and learns only collapse, as one module applied to
    every cell (4 inputs, 32 hidden units, 4 outputs; 292 parameters).
    \item \textbf{L2} learns elimination as a second shared module applied
    to every candidate. The harness supplies its nine inputs: whether the
    candidate's own cell holds this digit, whether it holds another digit,
    and whether each of its seven peer cells holds this digit (9 inputs,
    32 hidden units, 1 output; 353 parameters). The thresholded output
    feeds the collapse module of L1.
    \item \textbf{L3} learns a single operator that maps the whole board
    (64 candidate bits) to the cells confirmed in the next wave; an
    external loop writes them back. L3$^{*}$ is the same network trained
    on the full solution instead of the next wave.
    \item \textbf{L4} is a network with recurrent hidden connections that
    maps the givens to the full solution. L4-33k has 128 recurrent hidden
    units, which matches the parameter count of L5 at half its width. L4-ds
    is L4 with deep supervision: it is presented for 80 steps in five
    windows of 16, and the readout of window $k$ is trained on the cells
    known after $k$ waves, so that internal recurrence receives the same
    per-wave targets as L3.
    \item \textbf{L5} is a feedforward network that maps the givens to the
    full solution in one pass.
\end{itemize}
L1 and L2 are positive controls rather than tests of generalization: their
modules see only 15 and 95 distinct input patterns, all of which occur in
1,000 training puzzles, and they cannot see which grid they are working
on.

\subsection{Networks}
\label{sec:networks}
All learned spiking networks share one core: an input layer, a hidden
layer of 256 leaky integrate-and-fire (LIF) neurons (128 in L4-33k), and a
leaky-integrator readout. Before the spike nonlinearity, the hidden membrane potential is
\begin{equation}
    u_t = \beta\,(u_{t-1} - s_{t-1}) + W_{\mathrm{in}} x_t + b_{\mathrm{in}} + W_{\mathrm{rec}} s_{t-1},
    \qquad s_t = H(u_t - 1),
    \label{eq:lif}
\end{equation}
where $x_t$ is the input spike vector, $s_t$ the hidden spike vector,
$\beta=0.9$ the membrane decay, and $H$ the Heaviside step function. The
threshold is 1, the reset subtracts the emitted spike, and the gradient is
not propagated through the reset term. $W_{\mathrm{rec}}$ is present only
in L4, L4-33k and L4-ds. The readout potential
$v_t = \beta v_{t-1} + W_{\mathrm{out}} s_t + b_{\mathrm{out}}$ has no
threshold, and its average over the last 16 of 48 simulated steps is the
network output (a logit per output unit). In the backward pass the
derivative of $H$ is replaced by the fast-sigmoid surrogate
$1/(1+5|u-1|)^2$ \citep{neftci2019,zenke2018}. Each bit of the 64-bit
board is converted to a spike at every time step by an independent
Bernoulli draw with probability 0.80 for an active bit and 0.02 for an
inactive one. Weights start from the PyTorch default initialization,
after which $W_{\mathrm{in}}$ and $W_{\mathrm{out}}$ are multiplied by 3
and $W_{\mathrm{rec}}$ by 0.5. L3, L3$^{*}$ and L5 have 33,088 parameters,
L4 and L4-ds 98,624, and L4-33k 32,960.

Each learned level has a non-spiking counterpart of the same size. For
L3 to L5, including L3$^{*}$, L4-33k and L4-ds, it is a leaky rate network
with the same layers and readout, whose hidden state follows $h_t = \beta h_{t-1} +
(1-\beta)\,\mathrm{ReLU}(W_{\mathrm{in}} x_t + b_{\mathrm{in}} +
W_{\mathrm{rec}} h_{t-1})$. The main counterparts receive the clean binary
input; a second set, the noisy-input counterparts, receives the same
Bernoulli spike trains as the spiking networks. The two then differ in
the spiking nonlinearity, the reset and the surrogate gradient, and also
in the order of integration and nonlinearity: the LIF neuron sums its
input over time before thresholding it, whereas the rate network applies
the ReLU at every step and then averages. A third, post hoc counterpart
for L3 removes only the threshold and the reset: its hidden units
integrate the same spike trains as the LIF membrane,
$u_t = \beta u_{t-1} + W_{\mathrm{in}} x_t + b_{\mathrm{in}}$, and emit
$h_t = \mathrm{ReLU}(u_t)$ (the integrate-first counterpart).
For L1 the collapse module's counterpart is a logistic regression and for
L2 the elimination module's is a leaky rate network. The feedback
policies' counterpart is a multilayer perceptron (MLP) with the same layer
sizes.

\subsection{Supervised training and evaluation protocol}
L1 and L2 modules are trained on targets computed by the harness from the
boards that occur while solving the training puzzles. L3 is trained on
pairs of board and next-wave target, L3$^{*}$ on pairs of board and full
solution, and L4, L4-33k and L5 on pairs of givens and full solution. All
arms use binary cross-entropy on the output logits, the Adam optimizer
\citep{kingma2015}, batches of 256, and a fixed budget of 15,700 updates
with a cosine-decayed learning rate. The initial learning rate was chosen
separately for every arm of L3 to L5 and every network family from
$\{10^{-3}, 3\times10^{-3}, 10^{-2}, 3\times10^{-2}\}$. Each candidate was
trained with two seeds on a split whose validation puzzles come from 29
grids removed from the training grids, so that the choice does not reward
recall; the rate with the highest mean validation solve rate was kept,
and rates within 0.5 points of the best counted as ties and went to the
smaller rate. The held-out test grids played no part in the choice. The
chosen rates are listed in Table~\ref{tab:main}; spiking networks mostly
needed $10^{-2}$ (L3 $3\times10^{-3}$), non-spiking ones mostly $10^{-3}$
or $3\times10^{-3}$ (L3$^{*}$ $10^{-2}$), and $3\times10^{-2}$ made every
non-spiking network diverge. The integrate-first counterpart was tuned by
the same rule ($10^{-3}$). The small L1 and
L2 modules use $10^{-2}$, chosen on validation puzzles. Every 1,570
updates the arm is evaluated on a fixed random 600 of the validation
puzzles, whose depth mix follows the pool, and the checkpoint with the
highest validation solve rate is kept (for the L2 elimination module, the
highest candidate accuracy). Looped arms (L1, L2, L3, L3$^{*}$) write
back, in each pass, the most probable digit of every empty cell whose
probability exceeds a threshold chosen on the same validation puzzles
from 0.3 to 0.9, for at most 8 passes. L4 and L5 decode each cell as its
most probable digit, givens included; clamping the givens changes their
solve rates by at most 0.4 points. Seeds change the initialization and
the spike noise but not the order of minibatches, which is the same in
every seed.

\subsection{Feedback environment and training}
\label{sec:feedback-env}
In the stepwise feedback condition the policy sees the placed digits and
empty-cell flags (80 inputs) and picks one empty cell and one digit per
step (64 logits, filled cells masked, illegal digits not masked); the
environment writes the move onto the board. An episode ends with reward
$+1$ when the board is full and $-1$ when the chosen digit already appears
in a peer cell (an illegal move), when an empty cell has no legal digit
left (a dead end), or after 16 steps. The network is never told which
digit is correct. Each decision uses 16 simulated steps. The spiking
policy has 256 hidden LIF neurons (37,184 parameters). Policies are
trained with REINFORCE \citep{williams1992} and Adam at a constant
learning rate of $10^{-3}$: the return of step $k$ in an episode of $K$
steps is $\gamma^{K-1-k}R$ with $\gamma=0.9$, the baseline is the mean
return over all steps in the batch, and an entropy bonus of 0.01 is
added. Training runs for 30,000 updates of 64 episodes. Evaluation uses
the greedy action.

Seven variants test what limits this learner.
\begin{itemize}
    \item \textbf{Delay manipulation.} The training environment knows the
    solution and declares a dead end $k$ moves after the first move that
    disagrees with it ($k\in\{0,1,2,4\}$). Moves before the fatal one
    still share its discounted return, so the manipulation shortens the
    delay without removing credit ambiguity.
    \item \textbf{Answer oracle.} Every move is rewarded at once: $+1$ if it
    agrees with the solution, otherwise $-1$ and the episode ends; the
    return of a move is its own reward ($\gamma=0$). Correct moves that are
    not forced, that is, lucky guesses between two open candidates, are
    rewarded too.
    \item \textbf{Rule oracle.} As the answer oracle, but $+1$ only for a
    forced move (a cell with exactly one legal digit, filled with that
    digit); any other move is $-1$ and ends the episode. The reward then
    matches the fewest-legal-digits rule exactly.
    \item \textbf{Imitation.} The same policy network is trained by
    supervision to imitate the fewest-legal-digits rule: along the rule's
    trajectories through the training puzzles, the target is the uniform
    distribution over all forced (cell, digit) pairs.
    \item \textbf{Stepwise, normalized.} As the standard condition, with
    advantages standardized within each batch and 60,000 updates.
    \item \textbf{Stepwise, whole-episode reward.} The environment writes
    every move, including illegal ones, and never ends an episode early;
    when the board is full it returns $+1$ if there is no conflict and
    otherwise minus the fraction of the 56 peer pairs that conflict.
    \item \textbf{One-shot decisions.} The policy outputs the whole board in
    one decision (48 simulated steps, read over the last 16), with the same
    whole-board reward (FB-shot) or with a reward of $-1$ or $+1$ for each
    cell depending on whether it conflicts with a peer (FB-cell).
\end{itemize}
The last four conditions form a $2\times2$ design (stepwise or one-shot,
finer or coarser reward); all four standardize advantages and train for
60,000 updates. The factor levels are not identical across the two
decision types: the finer reward of the stepwise arms is a per-move
termination signal, that of the one-shot arms a per-cell conflict signal,
and the termination rules differ as well. All variants are tested in the
standard environment.

\subsection{Noise controls}
To ask whether spiking helps under input noise, we flipped a fraction
$p\in\{0, 2, 5, 10\}\%$ of the 64 bits of the board at test time, before
the Bernoulli spike encoding, for the L3 operator: (a) spiking, flips
redrawn at every time step; (b) spiking, one flip pattern held for all
steps; (c) MLP trained on clean input, one flipped input; (d) MLP trained
on single Bernoulli frames, one frame; (e) the same MLP given 16
independent noisy frames, with its output probabilities averaged over the
frames, and (e$'$), as a check, the same average over 48 frames, the
number of steps in one spiking presentation; (f) the clean-trained leaky rate
counterpart, flips redrawn at every step; and (g) the noisy-input leaky
rate counterpart, flips redrawn at every step. Two post hoc checks ask
whether integrating the input before the nonlinearity is enough: (h) the
integrate-first counterpart (Section~\ref{sec:networks}), flips redrawn at every step or
one flip pattern held; (i) the clean-trained MLP of (c), given the average
of the 48 noisy Bernoulli input frames instead of one input, and (i$'$)
the same average binarized at 0.5 before it enters the MLP. For (i) and
(i$'$), whose inputs differ from those the MLP was trained on, the
write-back threshold was chosen again on clean validation puzzles. The MLP learning rate was
chosen on validation puzzles from $\{10^{-3}, 3\times10^{-3}, 10^{-2},
3\times10^{-2}\}$ ($10^{-2}$ for both MLPs); at the default $10^{-3}$ the
clean MLP was badly under-trained. Conditions (a), (b), (f) and (g) use
the tuned L3 networks of the main ladder. Conditions (a) to (f) were
specified before they were run; (g), (h), (i), (i$'$), the 48-frame
average and the learning-rate selection were added afterwards, (h) to
(i$'$) with their reading criteria recorded before they were run.

\section{Experimental setup}
\label{sec:setup}

\subsection{Configurations compared}
Table~\ref{tab:arms} lists the arms. Main supervised arms of all three
families run with five seeds; data-efficiency, shallow-training and
robustness arms run with three. Stepwise feedback policies run with five
seeds and variants with three.

\begin{table}[t]
\caption{Arms (``post hoc'' = added after the pre-specified analysis).}
\label{tab:arms}
\centering
\resizebox{\textwidth}{!}{%
\begin{tabular}{llc}
\toprule
Arm & What varies & Seeds \\
\midrule
L1, L2, L3, L3$^{*}$, L4, L4-33k, L5 & encapsulation level (spiking) & 5 \\
L1--L5 counterparts (clean input) & non-spiking & 5 \\
L1--L5 at 1k, 4k, 16k puzzles & training set size & 3 \\
L3, L4, L5 shallow & trained on depths 1--2 only & 3 \\
FB-step, FB-step MLP; FB-delay $k$ & stepwise policy; fatal-move delay & 5; 3 \\
FB-stepwhole, FB-shot, FB-cell & interaction and reward granularity & 3 \\
Noise controls (a)--(f); (g) post hoc & L3 under input noise & 5 \\
\midrule
L4-ds; L3$^{*}$, L4-33k, L4-ds counterparts (post hoc) & per-wave targets & 5 \\
Noisy-input counterparts of L3 to L5 (post hoc) & input noise matched & 5 \\
L3, L3$^{*}$, L4, L4-33k, L5 with held-out validation grids (post hoc) & model selection & 3 \\
L3, L3$^{*}$, L4, L5 on class split and on type split (post hoc) & stricter holdouts & 3 \\
Answer oracle, rule oracle, imitation, FB-step normalized (post hoc) & credit and reward & 3 \\
Learning-rate selection, all arms of L3 to L5 in three families (post hoc) & learning rate & 2 \\
\bottomrule
\end{tabular}}
\end{table}

\subsection{Pre-specification and deviations}
The main study's design, six hypotheses and analysis plan were committed
to the project's version-controlled repository before any of its runs
(commits 1a58cd9 and de84f0b), and the split hash and all protocol changes
were recorded before the main batch (commit 1660603); the noise protocol
(cd80dc1) and the post hoc controls (1868990, ab3c14d) were likewise
recorded before they were run. The record is an internal git history
rather than a public registry; it becomes inspectable when the repository
is released. The deviations from the plan are these. (i) The first
version of the loss clipped probabilities to $[10^{-4}, 1-10^{-4}]$,
which zeroes the gradient of saturated outputs; after it stalled the L1
module in a learning gate, all arms use the loss on logits. (ii) The
learning rate of the L1 and L2 modules was chosen on validation puzzles,
as allowed. (iii) The plan named MLPs and a gated recurrent unit as
counterparts; we used leaky rate networks of the same size for L3 to L5 so
that the counterparts share the spiking networks' temporal readout.
(iv) The plan called for holding out grids by symmetry class; the main
split holds out 58 random grids, and the class and type splits were run
afterwards as checks. (v) The shallow-training arms of L1 and L2 were not
run. (vi) The plan named Wilson intervals; we report two-level bootstrap
intervals, which also include seed variance. (vii) The MLP learning rate
in the noise controls was re-selected on validation puzzles after the
pre-specified analysis. (viii) The noise condition (g) and the rows marked
post hoc in Table~\ref{tab:arms} were added after the pre-specified
analysis. (ix) The plan fixed one learning rate, $10^{-3}$, for all
networks of L3 to L5. Prompted by feedback on an earlier draft, a sweep showed it to be
too low for most of them, and every supervised arm of L3 to L5, with the
data-efficiency, shallow-training, robustness and noise analyses that
depend on it, was re-run at a learning rate chosen for that arm. The
selection rule and three criteria for reading the re-run results (whether
the loop and per-wave targets are both needed, whether the recall
interpretation stands, and whether the spiking operator beats every
non-spiking control under noise) were committed before the runs (commits
96d2ad0 and fc00463); all three criteria were met. (x) Later, the
integrate-first counterpart and the averaged-input MLP checks (h) to
(i$'$) were added, with their reading criteria committed before the runs
(commit a753fcb); against them the third criterion is no longer met
(Section~\ref{sec:spike-res}). Results at the common
rate are listed in Table~\ref{tab:untuned}. The feedback policies were not
re-tuned.

\subsection{Environment and evaluation}
All networks were implemented in PyTorch \citep{paszke2019}. The
supervised networks of L3 to L5 were trained on the GPU of a single
laptop (NVIDIA RTX 5090 Laptop, PyTorch 2.14.0 with CUDA 13.0), with each
training step captured as a CUDA graph; on networks without random input,
this path reproduces the loss trajectory of ordinary execution to four
decimals. L1, L2, the feedback policies and the MLP noise controls were
trained on the laptop's CPU (PyTorch 2.13.0). The GPU draws spikes from a
different random stream, so GPU and CPU runs with the same seed are not
identical.
The hand-wired solver was simulated in Brian~2
\citep{stimberg2019} with a 0.1\,ms time step. The primary metric is the
whole-puzzle solve rate by depth: a puzzle counts only if every cell is
correct. We report means over seeds with 95\% confidence intervals (CIs)
from a two-level bootstrap that resamples seeds and puzzles; comparisons
between arms resample the seeds of each arm independently and the puzzles
jointly. With three or five seeds the seed level of this bootstrap has few
distinct resamples, so per-seed results are listed in
Table~\ref{tab:perseed} in the Appendix. Per-seed exact McNemar tests
\citep{mcnemar1947} are included with the result files. We do not correct
for the roughly 30 comparisons; effects whose intervals come near zero
should be read as tentative. Training, analysis and figure scripts were
written with the assistance of Claude (Anthropic); every script was run
and its output checked by the authors, and all numbers reported here are
read from the saved result files.

\section{Results}
\label{sec:results}

\subsection{A hand-wired reference (L0)}
\label{sec:l0}
A hand-wired spiking solver built from the same candidate code (64
candidate neurons, 640 fixed inhibitory synapses, 64 disinhibition-gated
confirm neurons, with confirmed digits written back as drive) solves 200
of 200 stratified puzzles (40 per depth, from the same pool) with membrane
noise up to $\sigma=2$\,mV and integration windows of at least 100\,ms.
Shorter windows or stronger noise reduce the solve rate, and a longer
window recovers it (at $\sigma=8$\,mV, 68.0\% at 100\,ms and 97.5\% at
200\,ms). This establishes that some small spiking circuit can represent
the task.

\subsection{The ladder on held-out solution grids}
\label{sec:ladder-res}
\begin{figure}[t]
    \centering
    \includegraphics[width=\textwidth]{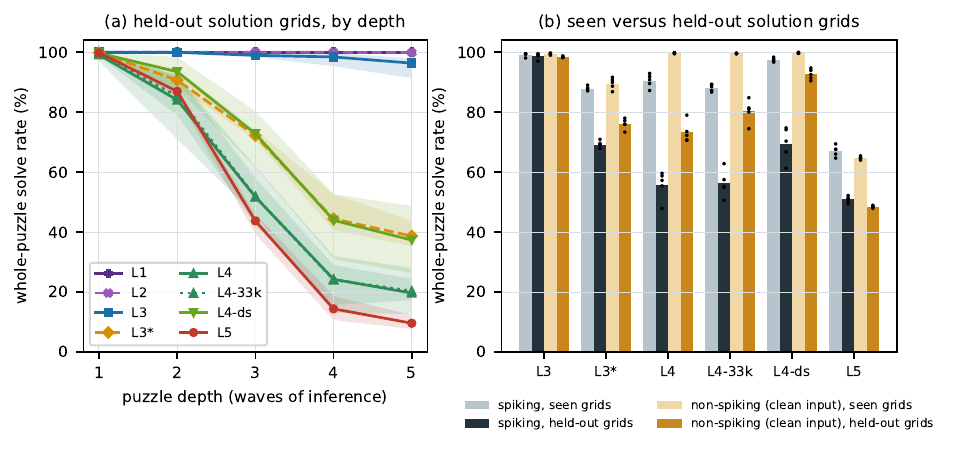}
    \caption{The encapsulation ladder on held-out solution grids. (a) Solve
    rate by depth for the spiking levels; bands show the range across
    seeds. L1 and L2 coincide at 100\%. (b) Overall solve rate of the same
    trained networks on new puzzles whose grids were seen in training
    (light bars) and on puzzles with held-out grids (dark bars), for spiking
    networks (gray) and clean-input non-spiking counterparts (orange); black
    dots are individual seeds. Each arm uses its own tuned learning rate.}
    \label{fig:ladderres}
\end{figure}

Table~\ref{tab:main} and Fig.~\ref{fig:ladderres} report the learned
levels. The positive controls behave as expected: L1 solves every test
puzzle and L2 fails on 3 of 7,500 puzzle-seed pairs. With traversal also
to be learned, the one-wave operator in an external loop (L3) solves
98.7\% [97.7, 99.4] of held-out-grid puzzles, from 100\% at depth 1 to
96.3\% at depth 5, only 1.2 points [0.5, 2.3] below L2. On the boards of
the canonical solution path, 92.5\% of its passes write exactly the cells
of that wave, yet deviations seldom involve a wrong digit. In a separate
evaluation with fresh spike trains (98.5\% solved), 1.5\% of test puzzles
ended with a wrong digit somewhere and none was left incomplete after the
8 allowed passes; allowing 16 passes changed nothing. The arms that learn a full-solution
target, or that lack the external loop, solve far fewer puzzles: 69.1\%
with the external loop (L3$^{*}$), 55.8\% with internal recurrence (L4),
69.4\% with internal recurrence and per-wave deep supervision (L4-ds),
56.2\% for L4-33k and 50.9\% for one pass (L5).

\begin{table}[t]
\caption{Solve rate (\%) on held-out grids by depth, overall with two-level bootstrap CI, and on seen grids (five seeds each). Lr: the initial learning rate chosen for the arm.}
\label{tab:main}
\centering
\resizebox{\textwidth}{!}{%
\begin{tabular}{llccccccc}
\toprule
Arm & Lr & 1 & 2 & 3 & 4 & 5 & Held-out [95\% CI] & Seen \\
\midrule
\multicolumn{9}{l}{\textit{Spiking}} \\
L1 & $10^{-2}$ & 100 & 100 & 100 & 100 & 100 & 100.0 [100.0, 100.0] & 100.0 \\
L2 & $10^{-2}$ & 100 & 100 & 100 & 99.9 & 99.9 & 100.0 [99.9, 100.0] & 100.0 \\
L3 & $3\times10^{-3}$ & 100 & 99.9 & 98.9 & 98.4 & 96.3 & 98.7 [97.7, 99.4] & 99.2 \\
L3$^{*}$ & $10^{-2}$ & 99.8 & 90.7 & 72.2 & 44.4 & 38.7 & 69.1 [67.1, 71.2] & 87.8 \\
L4 & $10^{-2}$ & 99.1 & 84.1 & 51.7 & 24.3 & 19.7 & 55.8 [51.3, 59.7] & 90.5 \\
L4-ds & $10^{-2}$ & 99.7 & 93.5 & 72.8 & 43.9 & 37.3 & 69.4 [64.3, 73.9] & 97.4 \\
L4-33k & $10^{-2}$ & 99.5 & 85.3 & 51.8 & 24.1 & 20.3 & 56.2 [52.3, 60.3] & 88.2 \\
L5 & $10^{-2}$ & 99.8 & 86.9 & 43.7 & 14.4 & 9.6 & 50.9 [48.6, 53.3] & 67.1 \\
\midrule
\multicolumn{9}{l}{\textit{Non-spiking, clean input}} \\
L3 & $10^{-3}$ & 100 & 99.9 & 99.0 & 98.4 & 95.0 & 98.5 [98.0, 98.9] & 99.4 \\
L3$^{*}$ & $10^{-2}$ & 100 & 99.5 & 83.3 & 51.3 & 46.3 & 76.1 [73.8, 78.2] & 89.5 \\
L4 & $3\times10^{-3}$ & 98.1 & 90.0 & 80.2 & 50.0 & 47.9 & 73.2 [70.2, 76.6] & 99.7 \\
L4-ds & $3\times10^{-3}$ & 100 & 100 & 98.3 & 88.1 & 76.7 & 92.6 [90.9, 94.2] & 99.8 \\
L4-33k & $3\times10^{-3}$ & 98.8 & 95.8 & 86.4 & 62.4 & 58.3 & 80.3 [76.9, 83.4] & 99.6 \\
L5 & $10^{-3}$ & 100 & 93.8 & 41.8 & 4.2 & 2.5 & 48.5 [46.0, 51.0] & 64.7 \\
\midrule
\multicolumn{9}{l}{\textit{Non-spiking, noisy input (the spike trains of the spiking networks)}} \\
L3 & $3\times10^{-3}$ & 100 & 99.9 & 98.1 & 96.7 & 91.7 & 97.3 [96.3, 98.1] & 98.3 \\
L3$^{*}$ & $10^{-2}$ & 99.9 & 92.7 & 58.2 & 31.6 & 28.6 & 62.2 [59.9, 64.5] & 81.3 \\
L4 & $3\times10^{-3}$ & 97.7 & 84.3 & 63.1 & 29.5 & 26.5 & 60.2 [56.7, 63.9] & 97.7 \\
L4-ds & $3\times10^{-3}$ & 99.6 & 95.1 & 85.9 & 57.5 & 43.5 & 76.3 [74.6, 78.1] & 98.7 \\
L4-33k & $3\times10^{-3}$ & 99.8 & 93.6 & 79.9 & 49.4 & 47.9 & 74.1 [71.4, 77.1] & 95.9 \\
L5 & $3\times10^{-3}$ & 100 & 86.1 & 31.2 & 5.7 & 3.7 & 45.3 [43.0, 47.7] & 61.4 \\
\bottomrule
\end{tabular}}
\end{table}

\runin{Loop and one-wave targets are both needed}
L3 differs from the full-solution arms in two ways at once: it has a
provided loop, and each pass is trained on one wave. The two controls
separate them (Table~\ref{tab:paired}). Keeping the loop but training each
pass on the full solution (L3$^{*}$) loses 29.6 points [27.4, 31.8].
Giving internal recurrence the same per-wave targets (L4-ds) gains 13.7
points over L4 [7.8, 19.6] but remains 29.3 points below L3 [24.8, 34.1],
level with L3$^{*}$ ($+0.3$ [$-4.6$, 4.9]). Each ingredient helps on its
own; only together do they generalize almost completely. The same holds
in both non-spiking families, with one difference in size: with clean
input, L4-ds comes within 5.8 points [4.2, 7.6] of L3 (92.6\% against
98.5\%), and the remaining gap lies at depths 4 and 5 (88.1\% and 76.7\%
against 98.4\% and 95.0\%). The pre-specified criterion for keeping the
claim that both ingredients are needed (L3 ahead of L4-ds by at least 5
points with an interval excluding zero, and ahead of L3$^{*}$ and L4) is
met in all three families, narrowly for clean-input networks. Intermediate
supervision therefore helps internal recurrence too, and its benefit is
largest when the loop that applies the supervised step is also provided.
Internal recurrence without intermediate targets is 4.9 points better
than one pass [0.5, 8.5], and at matched parameters as well (L4-33k
against L5: $+5.3$ [1.7, 9.3]); at matched parameters recurrence also
halves the width, so these data do not separate recurrence from width.
The ordering L3 $\gg$ L4 $>$ L5 holds in all three families.

\begin{table}[t]
\caption{Paired comparisons within each network family on held-out grids (difference in percentage points with two-level bootstrap 95\% CI).}
\label{tab:paired}
\centering
\resizebox{\textwidth}{!}{%
\begin{tabular}{lccc}
\toprule
Comparison & Spiking & Non-spiking, clean input & Non-spiking, noisy input \\
\midrule
L3 vs L3$^{*}$ & $+29.6$ [27.4, 31.8] & $+22.4$ [20.2, 24.6] & $+35.1$ [32.7, 37.5] \\
L3 vs L4-ds & $+29.3$ [24.8, 34.1] & $+5.8$ [4.2, 7.6] & $+21.0$ [19.1, 22.9] \\
L4-ds vs L4 & $+13.7$ [7.8, 19.6] & $+19.4$ [15.8, 22.6] & $+16.1$ [12.5, 19.3] \\
L4-ds vs L3$^{*}$ & $+0.3$ [$-4.6$, 4.9] & $+16.5$ [14.1, 19.0] & $+14.1$ [12.1, 16.1] \\
L3 vs L5 & $+47.8$ [45.3, 50.3] & $+50.0$ [47.6, 52.4] & $+51.9$ [49.5, 54.3] \\
L4 vs L5 & $+4.9$ [0.5, 8.5] & $+24.8$ [21.7, 28.4] & $+14.9$ [11.5, 18.3] \\
L4 vs L4-33k & $-0.4$ [$-5.9$, 4.5] & $-7.1$ [$-11.1$, $-2.6$] & $-13.9$ [$-17.9$, $-10.0$] \\
\bottomrule
\end{tabular}}
\end{table}

\runin{Recall of training material, and where it fails}
On new puzzles whose grids occurred in training, the full-solution arms do
far better than on held-out grids: L4 by 34.7 points (90.5\% against
55.8\%), L4-33k by 32.0, L4-ds by 28.0, L3$^{*}$ by 18.7 and L5 by 16.2.
For L3 the gap is 0.4 points [$-0.1$, 1.1] and for L1 and L2 it is zero,
by construction. The gaps are comparably large in both non-spiking families
(Table~\ref{tab:main}), except that the clean-input L4-ds loses only 7.2
points. The two test sets are matched in depth and number of givens. The
gaps do not come from near-duplicate puzzles: 67\% of the seen-grid test
puzzles have a training puzzle on the same grid whose givens are a strict
superset of theirs, but seen-grid accuracy is as high without such a
superset (L4 at depth 3: 96.0\% without, 93.3\% with, 51.7\% on held-out
grids). What the full-solution arms rely on is the grid itself. The
failures are also not spread evenly over held-out grids. On held-out
96-type grids the full-solution arms do well (depth 3: L4 91.3\%,
L3$^{*}$ 93.2\%, L4-ds 97.7\%, L5 81.6\%), and on held-out 192-type grids
they fail (depth 3: 41.4\%, 66.7\%, 66.3\% and 33.9\%), whereas L3 is the
same on both (99.0\% and 98.9\%). On the type split, which trains only on
192-type grids and tests only on 96-type grids at depths 1 to 3, all four
arms do well: L3 100.0\% [99.8, 100.0], L3$^{*}$ 97.8\% [96.6, 98.9], L4
92.9\% [90.0, 96.0] and L5 88.4\% [82.3, 93.6]. The recall failure of
full-solution networks is thus concentrated on the grid type that makes
up two thirds of all grids and alone admits deep puzzles. A likely reason
is that 96-type grids, being twice as self-symmetric, are more redundant:
fewer givens fix the whole grid, which would let a network complete them
from partial patterns, and the same redundancy may explain why they admit
only shallow puzzles. We have not tested this explanation.

\runin{Robustness to model selection and to the split}
Checkpoints and thresholds were selected on validation puzzles from
training grids, which could favor checkpoints that recall grids.
Selecting them instead on 29 validation grids drawn from the 230 training
grids and removed from training (the test set is unchanged), with
thresholds up to 0.98, changed the held-out solve rates, relative to
seeds 0 to 2 under the original selection, by $+0.6$ (L3), $-1.8$ (L5),
$-7.4$ (L3$^{*}$), $-7.5$ (L4) and $-8.3$ (L4-33k) points, and the
seen-minus-held-out gaps stayed large (L4: 37.1 points). On the class
split (Table~\ref{tab:robust}), with three held-out relabeling classes,
L3 keeps 98.5\% (a seen-minus-held-out gap of 1.0 point), whereas the
full-solution arms lose accuracy: L4 reaches 92.4\% on seen grids but
33.7\% on held-out classes. L3 is 64.8 points [60.4, 68.9] ahead of L4 and
48.2 points [44.1, 53.0] ahead of L3$^{*}$. At the untuned learning rate
of $10^{-3}$, L3 had reached only 72.0\% on held-out classes against
89.8\% on seen ones; with a tuned rate, what L3 learns is not tied to the
relabeling classes it has seen.

\begin{table}[t]
\caption{Robustness of the spiking arms: held-out validation grids, held-out relabeling classes, and the type split (solve rate \%, three seeds, tuned learning rates; the type split tests depths 1--3 only).}
\label{tab:robust}
\centering
\begin{tabular}{lccccc}
\toprule
& \multicolumn{2}{c}{Held-out validation grids} & \multicolumn{2}{c}{Class split} & Type split \\
Arm & Held-out & Seen & Held-out & Seen & Held-out \\
\midrule
L3 & 99.2 & 99.6 & 98.5 & 99.5 & 100.0 \\
L3$^{*}$ & 61.6 & 88.6 & 50.3 & 94.8 & 97.8 \\
L4 & 51.0 & 88.1 & 33.7 & 92.4 & 92.9 \\
L4-33k & 50.1 & 86.3 & -- & -- & -- \\
L5 & 48.5 & 68.9 & 44.8 & 72.0 & 88.4 \\
\bottomrule
\end{tabular}
\end{table}

\runin{Depth, number of empty cells and grid type}
Deeper puzzles have more empty cells (Table~\ref{tab:task}), and a
whole-puzzle score falls with the number of decisions even at constant
per-cell accuracy. Depth is also confounded with grid type, since only
192-type grids admit depths 4 and 5. A logistic regression of the
per-puzzle outcome on depth, number of empty cells and grid type, pooled
over five seeds with standard errors clustered by puzzle, gives depth
coefficients (per wave) of $-0.75$ (SE 0.16; L3), $-0.62$ (0.045; L4)
and $-0.92$ (0.064; L5); fitted on 192-type puzzles alone they are
$-0.72$ (0.16), $-0.60$ (0.045) and $-0.96$ (0.066). The empty-cell
coefficients (per cell) are $-0.17$ (0.11), $-0.34$ (0.031) and $-0.44$
(0.041), and the 192-type coefficient is $-0.39$ (0.66) for L3 but
$-2.42$ (0.17) for L4 and $-2.02$ (0.17) for L5. With L3 near ceiling its
coefficients are imprecise. Depth predicts failure beyond both the number
of cells to fill and the grid type, and grid type matters far more for
the full-solution arms than for L3.

\subsection{Data efficiency and learning order}
\begin{figure}[t]
    \centering
    \includegraphics[width=0.6\textwidth]{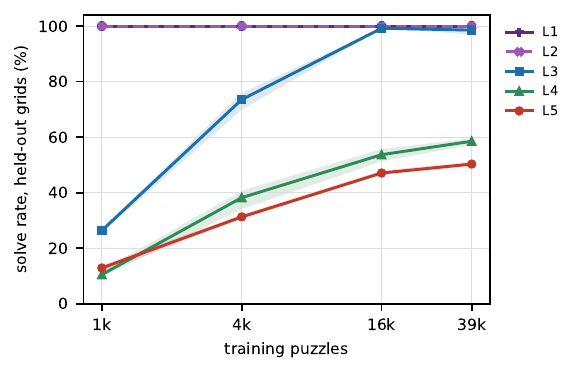}
    \caption{Data efficiency: solve rate on held-out grids against the
    number of training puzzles (logarithmic axis; three seeds, bands show
    the range). L1 and L2 coincide at 100\% at every size.}
    \label{fig:dataeff}
\end{figure}

Fig.~\ref{fig:dataeff} shows the solve rate on held-out grids after
training on 1,000 to 39,081 puzzles (three seeds, tuned learning rates;
the 39k point is the mean of seeds 0 to 2). L1 and L2 are at ceiling from
1,000 puzzles, as their input spaces are fully covered. L3 rises from
26.4\% to 99.2\% at 16,000 puzzles, and L4 and L5 from 10.5\% and 12.9\% to
53.7\% and 47.1\%. At 4,000 puzzles the order is L3 (73.5\%) $>$ L4
(38.2\%) $>$ L5 (31.3\%), so the provided loop with one-wave targets buys a
large gain in data efficiency, and internal recurrence a small one.

We also recorded, every 1,570 updates, the accuracy on validation puzzles
of the cells confirmed at each integration breadth. We had predicted that
breadth-1 deductions would be learned first. At the first checkpoint L3
was slightly more accurate on breadth-1 cells than on breadth-3 cells
(86.1\% against 83.1\%), but from the fourth checkpoint on, breadth-3
cells were at least as accurate as breadth-1 cells in L3, L4 and L5, the
reverse of the prediction, in the runs at the untuned learning rate. In
the tuned runs the pattern is mixed: at the first checkpoint L3 is more
accurate on breadth-3 than on breadth-1 cells (92.0\% against 89.2\%),
while L4 and L5 are slightly more accurate on breadth-1 cells (90.4\%
against 88.7\% and 90.5\% against 89.1\%). Because validation puzzles come
from training grids these curves also reflect recall, and we do not read
them as evidence about the order in which skills are acquired.

\subsection{Extrapolation from shallow training}
\begin{figure}[t]
    \centering
    \includegraphics[width=\textwidth]{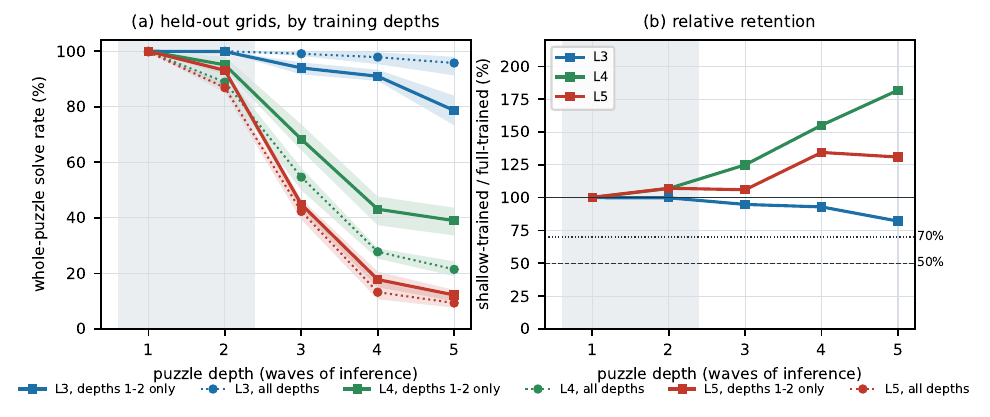}
    \caption{Extrapolation on held-out grids. (a) Solve rate by depth for
    networks trained on depths 1 and 2 only (solid, shaded region marks the
    trained depths) and on all depths (dotted), for L3 (blue), L4 (green)
    and L5 (red); three seeds, bands show the range. (b) Relative
    retention: the mean solve rate after shallow training divided by that
    after training on all depths; values above 100\% mean that shallow
    training generalized better. Horizontal lines mark the pre-specified
    thresholds of 70\% and 50\%.}
    \label{fig:extrap}
\end{figure}

Trained only on depths 1 and 2, L3 solves 91.0\% of depth-4 and 78.7\% of
depth-5 puzzles on held-out grids, 93\% and 82\% of what it reaches when
trained on all depths (Fig.~\ref{fig:extrap}; three seeds, tuned learning
rates). For L4 and L5 the comparison runs the other way. Trained only on
shallow puzzles, L4 is better overall on held-out grids (69.1\% against
58.5\% for the same seeds) and at depths 4 and 5 (43.1\% and 39.0\%
against 27.8\% and 21.4\%), and L5 is also better at depths 4 and 5
(17.8\% and 12.2\% against 13.2\% and 9.3\%). A
reading consistent with the recall results, which we have not tested, is
that deep training puzzles, with their few givens, reward recalling
training material, whereas shallow puzzles can be solved by local rules
that transfer. All shallow arms used shallow validation puzzles for model
selection.

\subsection{Spiking and non-spiking networks}
\label{sec:spike-res}
The ordering of the encapsulation levels holds for all three families;
the question is what spiking adds. With learning rates tuned, the spiking
L3 and L5 are as accurate as their clean-input counterparts or more so
(L3 $+0.3$ [$-0.8$, 1.1], L5 $+2.4$ [1.0, 3.9]), whereas the spiking
L3$^{*}$, also feedforward, is 6.9 points less accurate [4.6, 9.1]. All
three feedforward spiking networks are more accurate than their
noisy-input counterparts (L3 $+1.4$ [0.2, 2.5], L3$^{*}$ $+7.0$ [4.9,
9.0], L5 $+5.5$ [4.1, 7.0]), and the integrate-first counterpart of L3
solves 99.2\%, level with the spiking L3. The recurrent spiking networks are less accurate than both counterparts (L4
$-17.5$ [$-22.6$, $-13.0$] and $-4.5$ [$-9.6$, 0.0]; L4-ds $-23.2$ [$-28.1$,
$-18.5$] and $-6.9$ [$-11.6$, $-2.5$]; L4-33k $-24.1$ and $-17.9$). Part of
the difference between clean-input networks is the input noise itself,
since the noisy-input counterparts are less accurate than the clean-input
ones at every level; the rest, for recurrent networks, is consistent with
recurrent spiking networks being harder to train with surrogate
gradients, which we did not address beyond tuning the learning rate.
At the untuned common learning rate every spiking network had been less
accurate than its clean-input counterpart (Table~\ref{tab:untuned}); that
difference came mainly from the learning rate.

\begin{figure}[t]
    \centering
    \includegraphics[width=0.6\textwidth]{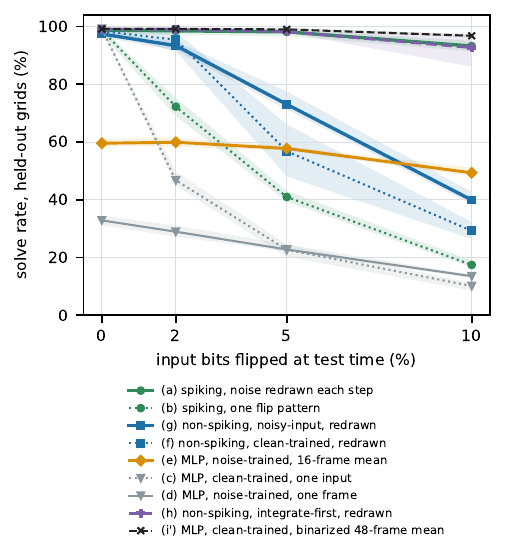}
    \caption{Input-noise robustness of the L3 operator on held-out grids
    (five seeds; bands show the range). Bits of the 64-bit board are
    flipped before spike encoding and redrawn at every time step unless
    noted. Green: spiking network, solid (a) with redrawn flips and dotted
    (b) with one flip pattern held over time. Blue: leaky rate network
    trained on the same spike trains (g, solid) or on clean input (f,
    dotted). Orange: MLP trained on single noisy frames and averaged over 16
    noisy frames at test (e). Gray: MLP trained on clean input and given one
    noisy input (c, dotted) or trained and tested on single noisy frames (d,
    solid). Purple, dashed: non-spiking network that integrates its input
    before the ReLU (h). Black, dashed: clean-trained MLP given the
    binarized mean of the 48 noisy input frames (i$'$). All learning rates
    were chosen on validation puzzles.}
    \label{fig:noise}
\end{figure}

Table~\ref{tab:noise} and Fig.~\ref{fig:noise} report the noise
controls. The clean spiking operator solves 98.8\% there rather than the
98.7\% of Table~\ref{tab:main}, because the spike trains are drawn anew.
Up to 5\% flips it barely changes (98.2\%), and at 10\% flips it still
solves 93.2\%, ahead of the noisy-input non-spiking network by 53.3 points
[50.6, 56.0] and of the MLP averaged over 16 frames, the strongest MLP
control, by 43.9 points [41.4, 46.5]; averaging over 48 frames did not
help the MLP. With one flip pattern held over time, the spiking operator
falls to 17.4\%, so its advantage depends on noise that varies in time.

These controls all apply their nonlinearity before averaging over time,
whereas the LIF neuron integrates its input first; redrawn flips that are
averaged before the nonlinearity only move the firing probabilities of
active and inactive bits from 0.80 and 0.02 to about 0.72 and 0.10. The
integrate-first counterpart (h), which keeps this order but has no
threshold and no reset, matches the spiking operator at every flip rate
(92.7\% at 10\%; difference $+0.5$ [$-3.2$, 4.7]) and collapses in the
same way when one flip pattern is held (15.4\%). A clean-trained MLP that
receives the average of the 48 noisy input frames (i) keeps 77.4\% at 10\%
flips, and with that average binarized at 0.5 (i$'$) it keeps 96.8\%, 3.5
points [2.1, 5.0] more than the spiking operator. The criterion recorded
before these checks, that threshold and reset add robustness beyond
integration (the spiking operator ahead of (h) by at least 5 points with
an interval above zero), is not met, and the spiking operator no longer
beats every non-spiking control. Its robustness to time-varying noise
comes from integrating the input before the nonlinearity, which the LIF
neuron does by construction, not from the spikes themselves.

\begin{table}[t]
\caption{Solve rate (\%) of the L3 operator on held-out grids under test-time input flips (five seeds; tuned learning rates). Flips are redrawn at every time step unless noted. Rows (h) to (i$'$) were added post hoc.}
\label{tab:noise}
\centering
\resizebox{\textwidth}{!}{%
\begin{tabular}{lcccc}
\toprule
Condition & 0\% & 2\% & 5\% & 10\% \\
\midrule
(a) Spiking & 98.8 & 98.5 & 98.2 & 93.2 \\
(b) Spiking, one flip pattern held & 98.8 & 72.3 & 41.0 & 17.4 \\
(c) MLP, clean-trained, one input & 99.2 & 46.8 & 22.7 & 10.1 \\
(d) MLP, noise-trained, one frame & 32.9 & 28.9 & 22.8 & 13.5 \\
(e) MLP, noise-trained, mean of 16 frames & 59.6 & 59.9 & 57.8 & 49.3 \\
(e$'$) MLP, noise-trained, mean of 48 frames & 59.0 & 58.3 & 55.3 & 45.6 \\
(f) Leaky rate network, clean-trained & 98.5 & 95.4 & 56.8 & 29.3 \\
(g) Leaky rate network, noisy-input & 97.4 & 93.4 & 73.1 & 39.9 \\
\midrule
(h) Integrate-first network, noisy-input & 99.1 & 99.0 & 98.2 & 92.7 \\
(h) Integrate-first network, one flip pattern held & 99.1 & 68.5 & 37.1 & 15.4 \\
(i) MLP, clean-trained, mean of 48 input frames & 95.7 & 93.4 & 90.0 & 77.4 \\
(i$'$) As (i), mean binarized at 0.5 & 99.2 & 99.2 & 99.0 & 96.8 \\
\bottomrule
\end{tabular}}
\end{table}

\runin{Activity and a cost proxy}
Hidden neurons of the tuned spiking networks fire on 12\% to 27\% of time
steps (L3 0.267, L4 0.121, L5 0.159 spikes per neuron per step). Per
48-step presentation this amounts to about 0.64 million synaptic events
for L3, 0.55 million for L5 and 0.90 million for L4, 19\% to 41\% of the
multiply-accumulate operations of a dense network run for the same 48
steps, but many times the roughly 33,000 of a single MLP forward pass. At this activity and
presentation length the spiking networks offer no operation-count
advantage over a one-step non-spiking network; we did not optimize them
for sparsity.

\subsection{Feedback only: what limits a stepwise learner}
\begin{figure}[t]
    \centering
    \includegraphics[width=\textwidth]{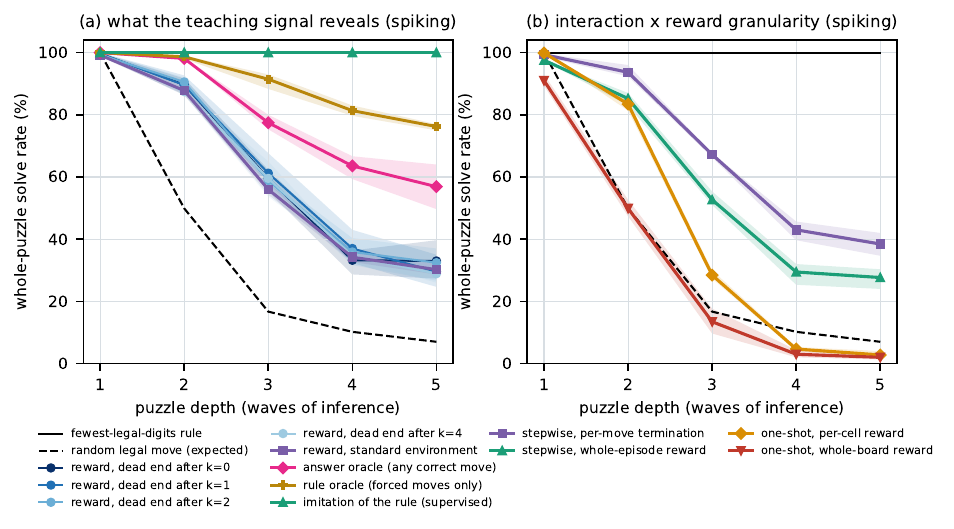}
    \caption{Learning from feedback, held-out grids (spiking policies; bands
    show the range across seeds). Black lines: the fewest-legal-digits rule
    (solid) and the exact expected solve rate of a random legal move
    (dashed). (a) Policies trained from reward with the dead end declared
    $k$ moves after the first fatal move (blue shades) or in the standard
    environment (purple); with the answer oracle (pink) or the rule oracle
    (gold) as reward; and by supervised imitation of the fewest-legal-digits
    rule (green). All are tested in the standard environment. (b) The
    $2\times2$ design with standardized advantages and 60,000 updates:
    stepwise policies with per-move termination (purple) or a whole-episode
    reward (green), and one-shot policies with per-cell (orange) or
    whole-board (red) reward.}
    \label{fig:feedback}
\end{figure}

Table~\ref{tab:feedback} and Fig.~\ref{fig:feedback} report the feedback
arms, all trained at the learning rate of Section~\ref{sec:feedback-env},
which was not re-tuned. The stepwise spiking policy learns the rules: on the held-out-grid
test set 8.8\% of its greedy episodes end in an illegal move, 30.4\% in a
dead end reached by legal moves, and 60.8\% [58.3, 63.1] in a solved
board. The MLP policy solves 64.2\%, 3.4 points more [1.4, 5.6]. Both do
much better on new puzzles with seen grids (83.8\% and 87.9\%), so
feedback-trained policies also rely on recalled material. Their learning
curves were still rising at the end of training, by about 3 points per
10,000 updates.

\begin{table}[t]
\caption{Feedback arms on held-out grids: solve rate (\%) by depth, overall with two-level bootstrap CI, and on seen grids.}
\label{tab:feedback}
\centering
\resizebox{\textwidth}{!}{%
\begin{tabular}{lccccccc}
\toprule
Arm & 1 & 2 & 3 & 4 & 5 & Held-out [95\% CI] & Seen \\
\midrule
Fewest-legal-digits rule & 100 & 100 & 100 & 100 & 100 & 100 & -- \\
Random legal move (expected) & 100 & 49.9 & 16.7 & 10.2 & 7.0 & 36.8 & -- \\
\midrule
FB-step, spiking & 99.4 & 87.1 & 55.5 & 34.0 & 28.1 & 60.8 [58.3, 63.1] & 83.8 \\
FB-step, MLP & 100 & 93.9 & 60.0 & 34.9 & 32.3 & 64.2 [62.2, 66.3] & 87.9 \\
FB-delay, $k=0$ & 99.7 & 89.7 & 59.6 & 33.2 & 33.0 & 63.0 [60.2, 65.9] & 85.2 \\
Answer oracle & 100 & 98.1 & 77.4 & 63.6 & 56.9 & 79.2 [76.4, 81.9] & 96.0 \\
Rule oracle & 100 & 98.7 & 91.4 & 81.3 & 76.2 & 89.5 [88.2, 90.8] & 95.8 \\
Imitation (supervised) & 100 & 100 & 100 & 100 & 100 & 100.0 [100.0, 100.0] & 100.0 \\
\midrule
FB-step, normalized & 99.3 & 93.7 & 67.2 & 43.0 & 38.4 & 68.3 [66.2, 70.5] & 90.3 \\
FB-stepwhole & 97.6 & 85.2 & 52.7 & 29.4 & 27.7 & 58.5 [55.5, 61.3] & 78.0 \\
FB-cell & 99.9 & 83.4 & 28.4 & 4.7 & 2.8 & 43.8 [41.5, 46.4] & 62.2 \\
FB-shot & 91.0 & 49.9 & 13.4 & 3.0 & 2.0 & 31.9 [29.6, 34.3] & 45.5 \\
\bottomrule
\end{tabular}}
\end{table}

\runin{Representation is not the limit}
Trained by supervised imitation of the fewest-legal-digits rule, the same
spiking policy network solves every held-out-grid test puzzle in every
seed. It can represent the rule and apply it to grids it has never seen.
What it does less well is learn the rule from reward.

\runin{Delay alone is not the limit; the match between reward and rule is}
The delay between a fatal move and its consequence grows with depth: in
the test set 61\% of wrong first moves at depth 5 lead to a dead end only
after further moves (Table~\ref{tab:task}), and in the policy's greedy
test episodes the share of fatal moves followed by further moves before
termination rises from 0\% at depth 2 to 74\% at depth 5. Declaring the
dead end at once ($k=0$) did not detectably help (63.0\% against 61.5\%
for the standard environment with the same seeds; difference 1.5 points
[$-1.3$, 4.5]), and the depth profiles for $k=0$ to 4 overlap
(Fig.~\ref{fig:feedback}a). Shortening the delay leaves the moves before
the fatal one sharing its discounted penalty, and already at depth 2,
where every fatal move is punished at once, the policy solves only 87\%.
Rewarding every move at once for agreeing with the answer removes this
ambiguity and raises the solve rate to 79.2\%, but it also rewards lucky
guesses between open candidates, which help only on grids that can be
recalled (96.0\% on seen grids). Rewarding only forced moves, so that the
reward matches the rule, raises the solve rate by a further 10.3 points
[7.3, 13.4] to 89.5\%, 28.0 points [25.9, 30.2] above the standard reward
and 10.5 points [9.1, 11.8] below imitation, with a seen-minus-held-out
gap of 6 points. With REINFORCE at this training budget, what limits the
learner is chiefly how well the reward identifies the moves that the rule
would make, first across moves and then among the alternatives at each
move, rather than the delay of consequences or the network's capacity.

\runin{Interaction and reward granularity}
In the $2\times2$ design (Fig.~\ref{fig:feedback}b), acting step by step on
a board that records each move adds 26.6 points [23.7, 29.5] with the
coarser reward and 24.5 points [22.3, 26.7] with the finer reward, and the
finer reward adds 9.8 points [6.9, 12.8] to stepwise learning and 12.0
points [10.1, 13.9] to one-shot learning. Both factors matter, and
stepwise interaction matters more. Because the finer reward is a
different signal in the two decision types, the one-shot and stepwise
conditions also differ in the number of decisions and simulated steps and
in how the policy conditions on its own earlier moves, and the one-shot
whole-board policy was still improving at the end of training, the size of
these effects is specific to this design.

\subsection{Earlier exploratory study and what changed}
\label{sec:e11}
The exploratory study used the same networks with three seeds, a test set
split by puzzle, and hyperparameters fixed while inspecting that test set.
Its headline results were that internal recurrence solved 91.3\% of test
puzzles against 55.4\% for one pass, and that a stepwise feedback policy
solved 81.3\%. These two numbers, for recurrence and for the feedback
policy, match the main study's results on seen grids
(90.5\% and 83.8\%) and far exceed those on held-out grids (55.8\% and
60.8\%); the earlier advantage of recurrence was mostly an advantage in
recalling training material. The exploratory study also suggested that
failures of the feedback learner reflected delayed consequences; the
delay manipulation, oracles and imitation locate the difficulty in the
match between reward and rule instead. One earlier observation held:
depth predicts failure beyond the number of empty cells.

\subsection{Pre-specified hypotheses}
Table~\ref{tab:prereg} lists the outcomes of the six hypotheses of the
main study, under the pre-specified protocol with one learning rate and
after the learning rates were tuned. Under the pre-specified protocol none
was fully supported, two were partly supported, and four were not. With
tuned learning rates, H1 is supported and H3 partly supported; the other
outcomes are unchanged, although for H4 the spiking advantage over the
MLP controls disappears against a non-spiking network that integrates its
input first.

\begin{table}[t]
\caption{Pre-specified hypotheses and outcomes, at the common learning rate of the plan and with tuned learning rates.}
\label{tab:prereg}
\centering
\small
\begin{tabular}{p{0.33\textwidth}p{0.30\textwidth}p{0.30\textwidth}}
\toprule
Hypothesis & Outcome, common rate & Outcome, tuned rates \\
\midrule
H1: at 4k puzzles, L1 $\geq$ L2 $\geq$ L3 $>$ L4 $>$ L5 & Partly: all but L4 $>$ L5 (24.8\% vs 27.1\%); L1, L2 at ceiling by construction & Supported (L3 73.5\%, L4 38.2\%, L5 31.3\%); L1, L2 at ceiling by construction \\
H2: breadth-1 deductions learned before breadth 2 and 3 at every level & Not supported: breadth-3 cells as accurate or more after the first checkpoint & Not supported: at the first checkpoint L3 more accurate on breadth-3 cells \\
H3: spiking within 5 points of non-spiking (L3--L5, clean input) & Not supported: non-spiking better by 6.0 to 13.8 points & Partly: L3 $+0.3$, L5 $+2.4$; L4 $-17.5$ \\
H4: under fair noise controls most of the spiking advantage vanishes (MLP averaged over 16 frames within 10 points of spiking) & Not supported: spiking ahead by 12.1 points at 10\% flips (tuned MLP) & Not supported against the averaged MLP (spiking ahead by 43.9 points); against the post hoc integrate-first network the advantage vanishes ($+0.5$) \\
H5: with $k=0$, feedback learning $\geq95\%$ at all depths; deep solve rate falls with $k$ & Not supported: 63.0\% at $k=0$; no trend with $k$ & Not supported (feedback arms not re-tuned) \\
H6: relative retention at depths 4--5 $\geq70\%$ for L1--L3, $<50\%$ for L5 & Partly: L3 69\% and 72\%; L5 above 100\%; L1, L2 not run & Partly: L3 93\% and 82\%; L5 above 100\%; L1, L2 not run \\
\bottomrule
\end{tabular}
\end{table}

\section{Discussion}
\label{sec:discussion}

\subsection{What must be built in}
Across the ladder, generalization to held-out grids depended on two
things provided together: an external loop that writes each result back,
and a target that asks each pass for one wave. With both, a spiking
operator learned elimination, collapse and traversal as one mapping and
generalized almost without loss, on held-out grids (0.4 points below seen
grids) and on held-out relabeling classes (1.0 point). With the loop but
full-solution targets, or with per-wave targets but only internal
recurrence, spiking networks improved over L4 but still recalled training
grids to a large extent and stayed about 30 points below L3. The
advantage of L3 is concentrated where recall does not help. On the type
split, which contains only 96-type grids and depths up to 3, L3 leads
L3$^{*}$ by 2.2 points; on the main split at depth 3 it leads by 5.8
points on 96-type grids and by 32.2 points on 192-type grids. Most of the
30-point overall advantage of the provided loop with one-wave targets
therefore comes from deep puzzles on the less-redundant grid type, where a
network cannot complete the solution from a recalled pattern.
Intermediate supervision is known to help neural networks learn
algorithms \citep{xu2020,velickovic2022,wies2023,shah2024}; here it helped
internal recurrence too, and helped most when the loop that applies the
supervised step was provided. How much the loop adds depends on the
network: in clean-input non-spiking networks, per-wave targets with
internal recurrence came within 6 points of the provided loop, in spiking
networks within 29. The unit traversal is the other large lever: providing
it (L2) turns the learning problem into a small, fully covered lookup and
removes the need to generalize at all. Within the limits of this task and
these learning rules, what has to be built in is the decomposition into
waves and, for spiking networks especially, the iteration that applies
it; for data efficiency, the structure of the units helps as well.

\subsection{What limits learning from feedback}
A spiking policy that represents the zero-lookahead rule perfectly when
taught by imitation learned it only partly from outcome reward (61\%).
Shortening the delay to the dead end did not help. Rewarding each move for
agreeing with the answer helped, but partly by rewarding lucky guesses
that transfer only to recalled grids; rewarding each move for following
the rule brought the learner to 89.5\%. Stepwise interaction with a
recording board mattered more than the granularity of the reward. For
closed-loop learners, the lesson is that a task can be easy to represent
and hard to learn from consequences, and that the difficulty lies in how
well the feedback singles out the right move, not merely in waiting for
the consequence.

\subsection{What spikes contribute}
Once each network had its own learning rate, the spiking L3 and L5 were
as accurate as their non-spiking counterparts on clean input, and all
feedforward spiking networks were more accurate than counterparts given
the same spike trains; much of the accuracy cost of spiking that the
untuned comparison showed came from the learning rate. The spiking
L3$^{*}$ and the recurrent spiking networks remained less accurate than
their clean-input counterparts. Under time-varying input noise the spiking
operator was far more robust than non-spiking networks that apply their
nonlinearity before averaging over time, but a non-spiking network that
integrates its input first, without threshold or reset, was just as
robust, and a conventional MLP given a time-averaged, binarized input was
more robust still. What protects the spiking operator is the order of
operations that the LIF neuron imposes, integration before the
nonlinearity, which any network can adopt; we found no robustness that
required spikes. The spiking networks were not sparse (0.12 to 0.27 spikes
per neuron per step), so any energy advantage would require explicit
sparsity.

\subsection{Failure modes and boundary conditions}
Three failure modes recur. Networks that output full solutions fail on
held-out grids of the larger grid type at depth 3 and beyond while
succeeding on seen grids. The learned one-wave operator rarely fails, and
when it does it is by an occasional wrong write in a deep puzzle, not by
running out of passes. Feedback-trained policies fail by legal moves that
lead into dead ends, and they do so even though the same network, taught
by imitation, makes no such moves.

\subsection{Limitations and external validity}
\label{sec:limits}
Several limitations bound these results. The networks are trained with
surrogate-gradient BPTT and REINFORCE, which are not local learning rules,
so the results describe what networks of this size reach under a strong,
non-local rule and do not predict what local rules such as e-prop
\citep{bellec2020} or three-factor plasticity \citep{fremaux2016} would
learn. The readout layer is a leaky integrator, not a spiking layer. The
task never requires inference beyond single candidates. There are only two
essentially different grids, so no split tests transfer to a truly new
solution structure at all depths; the type split covers depths 1 to 3
only. Only the learning rate was tuned, on a grid of four values; most
spiking arms chose $10^{-2}$, the second-largest value, so a finer grid
might still help them, and other settings (surrogate slope, input gain, presentation length)
might change the comparisons between families; the tuning was also added
after the pre-specified analysis, and results at the planned common rate
differ substantially (Table~\ref{tab:untuned}). The feedback policies
were not re-tuned, and their curves had not converged. Several variants
used three seeds, seeds shared their minibatch order, and the post hoc
controls were added after the pre-specified analysis. The comparison with
cultured neurons is a motivation only.

\subsection{Practical implications}
For researchers who build learned or neuromorphic solvers, the results
suggest four design checks. First, when the space of solution structures
is small, test on structures held out of training, and check the split
against the symmetries of the task; in our case a puzzle-level split
overstated generalization by up to about 38 points. Second, before interpreting
a difficulty-dependent failure as a limit of reasoning, test whether a
zero-lookahead rule already solves the task and whether the network can
learn it by imitation. Third, test an explanation by manipulating its
cause: the delay of consequences grew with depth and correlated with
failure, but removing it did not help. Fourth, when comparing spiking and
non-spiking networks, give both the same input, tune both, and match the
order of integration and nonlinearity: a common learning rate made every
spiking network look worse than its counterpart, tuning brought L3 and L5
level with theirs, and the apparent noise advantage of spiking vanished
against a non-spiking network that also integrates before its
nonlinearity.

\section{Conclusion}
\label{sec:conclusion}
We asked how much of a Sudoku solver must be built in before a small
spiking network can learn the rest, using $4\times4$ Sudoku as a testbed
whose difficulty is set by the exact number of inference waves and testing
on held-out solution grids, with learning rates tuned for every network.
A learned one-wave operator in a provided loop solved 98.7\% of
held-out-grid puzzles; the loop without one-wave targets (69.1\%) or
one-wave targets without the loop (69.4\%) generalized only partly, and
networks that output full solutions mainly recalled training grids,
failing most on the grid type that admits deep puzzles. The same ordering
held for non-spiking networks, where one-wave targets without the loop
came closer (92.6\% against 98.5\%). A spiking policy
that can represent the zero-lookahead rule perfectly learned it to 60.8\%
from outcome reward and to 89.5\% from a reward for rule-following moves;
shortening the delay of consequences did not help, and stepwise
interaction mattered more than reward granularity. The spiking operator
was as accurate as its non-spiking counterparts on clean input, and with
10\% of input bits flipped over time it still solved 93.2\% of puzzles;
a non-spiking network that integrates its input before the nonlinearity
was equally robust (92.7\%), so the robustness comes from that order of
operations rather than from spikes. Future work should test
tasks that require inference beyond single candidates, replace BPTT with
local learning rules, test whether networks can learn the iteration itself
and reuse a learned local operator at positions where it was never
trained, and find reward schemes that close the gap between learning from
consequences and learning by imitation.

\section*{Data availability}
\ifrepo
The code, puzzle pool, data splits, result files and design documents are
available at \url{\repourl}.
\else
The code, puzzle pool, data splits, result files and design documents will
be released in a public repository.
\fi
Because the data are generated by enumeration, no access restrictions
apply.

\section*{Use of generative AI}
During the preparation of this work the authors used Claude (Anthropic)
in order to check references, draft text and edit the English. After using this tool, the authors reviewed and
edited the content as needed and take full responsibility for the content
of the published article.

\appendix
\section{Per-seed results and results at the planned learning rate}
\setcounter{table}{0}
\renewcommand{\thetable}{A.\arabic{table}}
\begin{table}[H]
\caption{Per-seed overall solve rates (\%) on held-out grids (tuned learning rates for L3 to L5).}
\label{tab:perseed}
\centering
\small
\begin{tabular}{ll}
\toprule
Arm & Seeds 0, 1, 2, 3, 4 \\
\midrule
L1 & 100.0, 100.0, 100.0, 100.0, 100.0 \\
L2 & 99.9, 100.0, 100.0, 99.9, 100.0 \\
L3 & 99.2, 99.5, 97.0, 99.2, 98.7 \\
L3$^{*}$ & 70.9, 67.9, 68.0, 69.6, 69.3 \\
L4 & 59.6, 58.7, 57.3, 55.4, 47.9 \\
L4-33k & 62.8, 57.5, 54.9, 55.2, 50.6 \\
L4-ds & 74.1, 66.7, 74.7, 61.3, 70.3 \\
L5 & 51.2, 50.3, 49.4, 52.2, 51.3 \\
L3 clean-input counterpart & 98.5, 98.3, 98.7, 98.1, 98.6 \\
L3$^{*}$ clean-input counterpart & 76.0, 75.9, 77.1, 78.0, 73.3 \\
L4 clean-input counterpart & 79.0, 73.5, 72.3, 70.6, 70.7 \\
L4-33k clean-input counterpart & 74.5, 81.0, 80.1, 84.8, 81.4 \\
L4-ds clean-input counterpart & 94.7, 93.9, 91.4, 90.5, 92.5 \\
L5 clean-input counterpart & 48.5, 47.9, 48.5, 48.9, 48.5 \\
L3 noisy-input counterpart & 97.7, 96.1, 98.2, 96.6, 97.8 \\
L3$^{*}$ noisy-input counterpart & 61.1, 63.3, 60.9, 62.0, 63.7 \\
L4 noisy-input counterpart & 58.5, 56.0, 62.6, 58.5, 65.6 \\
L4-33k noisy-input counterpart & 74.2, 79.0, 72.3, 71.8, 73.3 \\
L4-ds noisy-input counterpart & 75.4, 77.1, 76.7, 75.8, 76.6 \\
L5 noisy-input counterpart & 45.2, 46.1, 45.1, 44.9, 45.3 \\
FB-step spiking & 61.5, 61.7, 61.3, 57.8, 61.7 \\
FB-step MLP & 65.3, 64.3, 63.8, 63.7, 64.1 \\
FB-delay $k=0$ & 62.2, 61.4, 65.5 \\
Answer oracle & 79.1, 76.8, 81.7 \\
Rule oracle & 90.1, 89.2, 89.3 \\
Imitation & 100.0, 100.0, 100.0 \\
FB-step normalized & 69.3, 68.3, 67.4 \\
FB-stepwhole & 60.2, 59.1, 56.3 \\
FB-cell & 43.8, 44.1, 43.6 \\
FB-shot & 31.3, 33.1, 31.3 \\
\bottomrule
\end{tabular}
\end{table}

\subsection*{Results at the common learning rate of the plan}
Table~\ref{tab:untuned} lists the supervised arms of L3 to L5 trained, as
pre-specified, at a common initial learning rate of $10^{-3}$ on the
laptop CPU. Two clean-input counterparts were run with one seed only, as
part of the learning-rate sweep.
\begin{table}[H]
\caption{Solve rate (\%) on held-out grids at the common learning rate $10^{-3}$, by depth and overall with two-level bootstrap CI.}
\label{tab:untuned}
\centering
\small
\begin{tabular}{lccccccc}
\toprule
Arm & Seeds & 1 & 2 & 3 & 4 & 5 & Overall [95\% CI] \\
\midrule
L3 & 5 & 100 & 97.9 & 87.3 & 78.3 & 59.2 & 84.5 [82.2, 86.7] \\
L3$^{*}$ & 5 & 96.5 & 81.1 & 44.5 & 19.1 & 16.1 & 51.5 [49.2, 53.7] \\
L4 & 5 & 96.1 & 71.7 & 41.5 & 14.3 & 11.7 & 47.1 [44.3, 49.8] \\
L4-ds & 3 & 96.6 & 73.9 & 43.9 & 18.2 & 17.9 & 50.1 [47.6, 52.6] \\
L4-33k & 5 & 91.9 & 66.7 & 30.8 & 9.9 & 7.7 & 41.4 [36.4, 45.3] \\
L5 & 5 & 99.5 & 75.1 & 25.9 & 7.2 & 4.3 & 42.4 [40.2, 44.8] \\
\midrule
L3, clean input & 5 & 100 & 99.7 & 98.9 & 98.3 & 94.9 & 98.4 [97.8, 98.9] \\
L3$^{*}$, clean input & 1 & 100 & 97.7 & 65.3 & 24.0 & 19.7 & 61.3 \\
L4, clean input & 5 & 96.9 & 80.2 & 59.6 & 24.3 & 24.3 & 57.1 [54.2, 60.1] \\
L4-ds, clean input & 1 & 100 & 98.7 & 87.3 & 57.0 & 49.3 & 78.5 \\
L5, clean input & 5 & 100 & 93.8 & 41.5 & 4.5 & 2.1 & 48.4 [46.0, 51.0] \\
\midrule
L3, noisy input & 5 & 100 & 95.9 & 76.3 & 61.0 & 41.9 & 75.0 [73.1, 77.0] \\
L4, noisy input & 3 & 99.6 & 86.0 & 62.0 & 28.4 & 25.2 & 60.2 [57.7, 62.8] \\
L4-ds, noisy input & 3 & 99.3 & 91.0 & 68.2 & 35.6 & 30.7 & 65.0 [61.6, 68.0] \\
L5, noisy input & 3 & 99.4 & 75.7 & 22.9 & 5.9 & 4.0 & 41.6 [39.2, 44.0] \\
\bottomrule
\end{tabular}
\end{table}

\bibliographystyle{apacite}
\bibliography{references}

\begin{thebibliography}{}

\bibitem [\protect \citeauthoryear {%
Amos%
\ \BBA {} Kolter%
}{%
Amos%
\ \BBA {} Kolter%
}{%
{\protect \APACyear {2017}}%
}]{%
amos2017}
\APACinsertmetastar {%
amos2017}%
\begin{APACrefauthors}%
Amos, B.%
\BCBT {}\ \BBA {} Kolter, J\BPBI Z.%
\end{APACrefauthors}%
\unskip\
\newblock
\APACrefYearMonthDay{2017}{}{}.
\newblock
{\BBOQ}\APACrefatitle {{OptNet}: Differentiable Optimization as a Layer in
  Neural Networks} {{OptNet}: Differentiable optimization as a layer in neural
  networks}.{\BBCQ}
\newblock
\BIn{} \APACrefbtitle {Proceedings of the 34th International Conference on
  Machine Learning} {Proceedings of the 34th international conference on
  machine learning}\ (\BVOL~70, \BPGS\ 136--145).
\PrintBackRefs{\CurrentBib}

\bibitem [\protect \citeauthoryear {%
Balci%
\ \protect \BOthers {.}}{%
Balci%
\ \protect \BOthers {.}}{%
{\protect \APACyear {2023}}%
}]{%
balci2023}
\APACinsertmetastar {%
balci2023}%
\begin{APACrefauthors}%
Balci, F.%
, {Ben Hamed}, S.%
, Boraud, T.%
, Bouret, S.%
, Brochier, T.%
, Brun, C.%
\BDBL {}Zimmermann, J.%
\end{APACrefauthors}%
\unskip\
\newblock
\APACrefYearMonthDay{2023}{}{}.
\newblock
{\BBOQ}\APACrefatitle {A Response to Claims of Emergent Intelligence and
  Sentience in a Dish} {A response to claims of emergent intelligence and
  sentience in a dish}.{\BBCQ}
\newblock
\APACjournalVolNumPages{Neuron}{111}{5}{604--605}.
\newblock
\begin{APACrefDOI} \doi{10.1016/j.neuron.2023.02.009} \end{APACrefDOI}
\PrintBackRefs{\CurrentBib}

\bibitem [\protect \citeauthoryear {%
Bansal%
\ \protect \BOthers {.}}{%
Bansal%
\ \protect \BOthers {.}}{%
{\protect \APACyear {2022}}%
}]{%
bansal2022}
\APACinsertmetastar {%
bansal2022}%
\begin{APACrefauthors}%
Bansal, A.%
, Schwarzschild, A.%
, Borgnia, E.%
, Emam, Z.%
, Huang, F.%
, Goldblum, M.%
\BCBL {}\ \BBA {} Goldstein, T.%
\end{APACrefauthors}%
\unskip\
\newblock
\APACrefYearMonthDay{2022}{}{}.
\newblock
{\BBOQ}\APACrefatitle {End-to-End Algorithm Synthesis with Recurrent Networks:
  Extrapolation without Overthinking} {End-to-end algorithm synthesis with
  recurrent networks: Extrapolation without overthinking}.{\BBCQ}
\newblock
\BIn{} \APACrefbtitle {Advances in Neural Information Processing Systems}
  {Advances in neural information processing systems}\ (\BVOL~35, \BPGS\
  20232--20242).
\PrintBackRefs{\CurrentBib}

\bibitem [\protect \citeauthoryear {%
Bellec%
\ \protect \BOthers {.}}{%
Bellec%
\ \protect \BOthers {.}}{%
{\protect \APACyear {2020}}%
}]{%
bellec2020}
\APACinsertmetastar {%
bellec2020}%
\begin{APACrefauthors}%
Bellec, G.%
, Scherr, F.%
, Subramoney, A.%
, Hajek, E.%
, Salaj, D.%
, Legenstein, R.%
\BCBL {}\ \BBA {} Maass, W.%
\end{APACrefauthors}%
\unskip\
\newblock
\APACrefYearMonthDay{2020}{}{}.
\newblock
{\BBOQ}\APACrefatitle {A Solution to the Learning Dilemma for Recurrent
  Networks of Spiking Neurons} {A solution to the learning dilemma for
  recurrent networks of spiking neurons}.{\BBCQ}
\newblock
\APACjournalVolNumPages{Nature Communications}{11}{}{3625}.
\newblock
\begin{APACrefDOI} \doi{10.1038/s41467-020-17236-y} \end{APACrefDOI}
\PrintBackRefs{\CurrentBib}

\bibitem [\protect \citeauthoryear {%
Buesing%
, Bill%
, Nessler%
\BCBL {}\ \BBA {} Maass%
}{%
Buesing%
\ \protect \BOthers {.}}{%
{\protect \APACyear {2011}}%
}]{%
buesing2011}
\APACinsertmetastar {%
buesing2011}%
\begin{APACrefauthors}%
Buesing, L.%
, Bill, J.%
, Nessler, B.%
\BCBL {}\ \BBA {} Maass, W.%
\end{APACrefauthors}%
\unskip\
\newblock
\APACrefYearMonthDay{2011}{}{}.
\newblock
{\BBOQ}\APACrefatitle {Neural Dynamics as Sampling: A Model for Stochastic
  Computation in Recurrent Networks of Spiking Neurons} {Neural dynamics as
  sampling: A model for stochastic computation in recurrent networks of spiking
  neurons}.{\BBCQ}
\newblock
\APACjournalVolNumPages{PLoS Computational Biology}{7}{11}{e1002211}.
\newblock
\begin{APACrefDOI} \doi{10.1371/journal.pcbi.1002211} \end{APACrefDOI}
\PrintBackRefs{\CurrentBib}

\bibitem [\protect \citeauthoryear {%
Chang%
, Flokas%
, Lipson%
\BCBL {}\ \BBA {} Spranger%
}{%
Chang%
\ \protect \BOthers {.}}{%
{\protect \APACyear {2020}}%
}]{%
chang2020}
\APACinsertmetastar {%
chang2020}%
\begin{APACrefauthors}%
Chang, O.%
, Flokas, L.%
, Lipson, H.%
\BCBL {}\ \BBA {} Spranger, M.%
\end{APACrefauthors}%
\unskip\
\newblock
\APACrefYearMonthDay{2020}{}{}.
\newblock
{\BBOQ}\APACrefatitle {Assessing {SATNet}'s Ability to Solve the Symbol
  Grounding Problem} {Assessing {SATNet}'s ability to solve the symbol
  grounding problem}.{\BBCQ}
\newblock
\BIn{} \APACrefbtitle {Advances in Neural Information Processing Systems}
  {Advances in neural information processing systems}\ (\BVOL~33, \BPGS\
  1428--1439).
\PrintBackRefs{\CurrentBib}

\bibitem [\protect \citeauthoryear {%
Cramer%
, Stradmann%
, Schemmel%
\BCBL {}\ \BBA {} Zenke%
}{%
Cramer%
\ \protect \BOthers {.}}{%
{\protect \APACyear {2022}}%
}]{%
cramer2022}
\APACinsertmetastar {%
cramer2022}%
\begin{APACrefauthors}%
Cramer, B.%
, Stradmann, Y.%
, Schemmel, J.%
\BCBL {}\ \BBA {} Zenke, F.%
\end{APACrefauthors}%
\unskip\
\newblock
\APACrefYearMonthDay{2022}{}{}.
\newblock
{\BBOQ}\APACrefatitle {The {Heidelberg} Spiking Data Sets for the Systematic
  Evaluation of Spiking Neural Networks} {The {Heidelberg} spiking data sets
  for the systematic evaluation of spiking neural networks}.{\BBCQ}
\newblock
\APACjournalVolNumPages{IEEE Transactions on Neural Networks and Learning
  Systems}{33}{7}{2744--2757}.
\newblock
\begin{APACrefDOI} \doi{10.1109/TNNLS.2020.3044364} \end{APACrefDOI}
\PrintBackRefs{\CurrentBib}

\bibitem [\protect \citeauthoryear {%
{Fonseca Guerra}%
\ \BBA {} Furber%
}{%
{Fonseca Guerra}%
\ \BBA {} Furber%
}{%
{\protect \APACyear {2017}}%
}]{%
fonsecaguerra2017}
\APACinsertmetastar {%
fonsecaguerra2017}%
\begin{APACrefauthors}%
{Fonseca Guerra}, G\BPBI A.%
\BCBT {}\ \BBA {} Furber, S\BPBI B.%
\end{APACrefauthors}%
\unskip\
\newblock
\APACrefYearMonthDay{2017}{}{}.
\newblock
{\BBOQ}\APACrefatitle {Using Stochastic Spiking Neural Networks on {SpiNNaker}
  to Solve Constraint Satisfaction Problems} {Using stochastic spiking neural
  networks on {SpiNNaker} to solve constraint satisfaction problems}.{\BBCQ}
\newblock
\APACjournalVolNumPages{Frontiers in Neuroscience}{11}{}{714}.
\newblock
\begin{APACrefDOI} \doi{10.3389/fnins.2017.00714} \end{APACrefDOI}
\PrintBackRefs{\CurrentBib}

\bibitem [\protect \citeauthoryear {%
Fr{\'e}maux%
\ \BBA {} Gerstner%
}{%
Fr{\'e}maux%
\ \BBA {} Gerstner%
}{%
{\protect \APACyear {2016}}%
}]{%
fremaux2016}
\APACinsertmetastar {%
fremaux2016}%
\begin{APACrefauthors}%
Fr{\'e}maux, N.%
\BCBT {}\ \BBA {} Gerstner, W.%
\end{APACrefauthors}%
\unskip\
\newblock
\APACrefYearMonthDay{2016}{}{}.
\newblock
{\BBOQ}\APACrefatitle {Neuromodulated Spike-Timing-Dependent Plasticity, and
  Theory of Three-Factor Learning Rules} {Neuromodulated spike-timing-dependent
  plasticity, and theory of three-factor learning rules}.{\BBCQ}
\newblock
\APACjournalVolNumPages{Frontiers in Neural Circuits}{9}{}{85}.
\newblock
\begin{APACrefDOI} \doi{10.3389/fncir.2015.00085} \end{APACrefDOI}
\PrintBackRefs{\CurrentBib}

\bibitem [\protect \citeauthoryear {%
Graves%
}{%
Graves%
}{%
{\protect \APACyear {2016}}%
}]{%
graves2016}
\APACinsertmetastar {%
graves2016}%
\begin{APACrefauthors}%
Graves, A.%
\end{APACrefauthors}%
\unskip\
\newblock
\APACrefYearMonthDay{2016}{}{}.
\newblock
{\BBOQ}\APACrefatitle {Adaptive Computation Time for Recurrent Neural Networks}
  {Adaptive computation time for recurrent neural networks}.{\BBCQ}
\newblock
\APACjournalVolNumPages{arXiv preprint arXiv:1603.08983}{}{}{}.
\newblock
\begin{APACrefDOI} \doi{10.48550/arXiv.1603.08983} \end{APACrefDOI}
\PrintBackRefs{\CurrentBib}

\bibitem [\protect \citeauthoryear {%
Habenschuss%
, Jonke%
\BCBL {}\ \BBA {} Maass%
}{%
Habenschuss%
\ \protect \BOthers {.}}{%
{\protect \APACyear {2013}}%
}]{%
habenschuss2013}
\APACinsertmetastar {%
habenschuss2013}%
\begin{APACrefauthors}%
Habenschuss, S.%
, Jonke, Z.%
\BCBL {}\ \BBA {} Maass, W.%
\end{APACrefauthors}%
\unskip\
\newblock
\APACrefYearMonthDay{2013}{}{}.
\newblock
{\BBOQ}\APACrefatitle {Stochastic Computations in Cortical Microcircuit Models}
  {Stochastic computations in cortical microcircuit models}.{\BBCQ}
\newblock
\APACjournalVolNumPages{PLoS Computational Biology}{9}{11}{e1003311}.
\newblock
\begin{APACrefDOI} \doi{10.1371/journal.pcbi.1003311} \end{APACrefDOI}
\PrintBackRefs{\CurrentBib}

\bibitem [\protect \citeauthoryear {%
Izhikevich%
}{%
Izhikevich%
}{%
{\protect \APACyear {2007}}%
}]{%
izhikevich2007}
\APACinsertmetastar {%
izhikevich2007}%
\begin{APACrefauthors}%
Izhikevich, E\BPBI M.%
\end{APACrefauthors}%
\unskip\
\newblock
\APACrefYearMonthDay{2007}{}{}.
\newblock
{\BBOQ}\APACrefatitle {Solving the Distal Reward Problem through Linkage of
  {STDP} and Dopamine Signaling} {Solving the distal reward problem through
  linkage of {STDP} and dopamine signaling}.{\BBCQ}
\newblock
\APACjournalVolNumPages{Cerebral Cortex}{17}{10}{2443--2452}.
\newblock
\begin{APACrefDOI} \doi{10.1093/cercor/bhl152} \end{APACrefDOI}
\PrintBackRefs{\CurrentBib}

\bibitem [\protect \citeauthoryear {%
Jolicoeur-Martineau%
}{%
Jolicoeur-Martineau%
}{%
{\protect \APACyear {2025}}%
}]{%
jolicoeur2025}
\APACinsertmetastar {%
jolicoeur2025}%
\begin{APACrefauthors}%
Jolicoeur-Martineau, A.%
\end{APACrefauthors}%
\unskip\
\newblock
\APACrefYearMonthDay{2025}{}{}.
\newblock
{\BBOQ}\APACrefatitle {Less is More: Recursive Reasoning with Tiny Networks}
  {Less is more: Recursive reasoning with tiny networks}.{\BBCQ}
\newblock
\APACjournalVolNumPages{arXiv preprint arXiv:2510.04871}{}{}{}.
\newblock
\begin{APACrefDOI} \doi{10.48550/arXiv.2510.04871} \end{APACrefDOI}
\PrintBackRefs{\CurrentBib}

\bibitem [\protect \citeauthoryear {%
Jonke%
, Habenschuss%
\BCBL {}\ \BBA {} Maass%
}{%
Jonke%
\ \protect \BOthers {.}}{%
{\protect \APACyear {2016}}%
}]{%
jonke2016}
\APACinsertmetastar {%
jonke2016}%
\begin{APACrefauthors}%
Jonke, Z.%
, Habenschuss, S.%
\BCBL {}\ \BBA {} Maass, W.%
\end{APACrefauthors}%
\unskip\
\newblock
\APACrefYearMonthDay{2016}{}{}.
\newblock
{\BBOQ}\APACrefatitle {Solving Constraint Satisfaction Problems with Networks
  of Spiking Neurons} {Solving constraint satisfaction problems with networks
  of spiking neurons}.{\BBCQ}
\newblock
\APACjournalVolNumPages{Frontiers in Neuroscience}{10}{}{118}.
\newblock
\begin{APACrefDOI} \doi{10.3389/fnins.2016.00118} \end{APACrefDOI}
\PrintBackRefs{\CurrentBib}

\bibitem [\protect \citeauthoryear {%
Kagan%
\ \protect \BOthers {.}}{%
Kagan%
\ \protect \BOthers {.}}{%
{\protect \APACyear {2022}}%
}]{%
kagan2022}
\APACinsertmetastar {%
kagan2022}%
\begin{APACrefauthors}%
Kagan, B\BPBI J.%
, Kitchen, A\BPBI C.%
, Tran, N\BPBI T.%
, Habibollahi, F.%
, Khajehnejad, M.%
, Parker, B\BPBI J.%
\BDBL {}Friston, K\BPBI J.%
\end{APACrefauthors}%
\unskip\
\newblock
\APACrefYearMonthDay{2022}{}{}.
\newblock
{\BBOQ}\APACrefatitle {In Vitro Neurons Learn and Exhibit Sentience when
  Embodied in a Simulated Game-World} {In vitro neurons learn and exhibit
  sentience when embodied in a simulated game-world}.{\BBCQ}
\newblock
\APACjournalVolNumPages{Neuron}{110}{23}{3952--3969.e8}.
\newblock
\begin{APACrefDOI} \doi{10.1016/j.neuron.2022.09.001} \end{APACrefDOI}
\PrintBackRefs{\CurrentBib}

\bibitem [\protect \citeauthoryear {%
Kingma%
\ \BBA {} Ba%
}{%
Kingma%
\ \BBA {} Ba%
}{%
{\protect \APACyear {2015}}%
}]{%
kingma2015}
\APACinsertmetastar {%
kingma2015}%
\begin{APACrefauthors}%
Kingma, D\BPBI P.%
\BCBT {}\ \BBA {} Ba, J.%
\end{APACrefauthors}%
\unskip\
\newblock
\APACrefYearMonthDay{2015}{}{}.
\newblock
{\BBOQ}\APACrefatitle {Adam: A Method for Stochastic Optimization} {Adam: A
  method for stochastic optimization}.{\BBCQ}
\newblock
\BIn{} \APACrefbtitle {Proceedings of the 3rd International Conference on
  Learning Representations ({ICLR}).} {Proceedings of the 3rd international
  conference on learning representations ({ICLR}).}
\newblock
\APACaddressPublisher{San Diego, CA, USA}{}.
\PrintBackRefs{\CurrentBib}

\bibitem [\protect \citeauthoryear {%
McNemar%
}{%
McNemar%
}{%
{\protect \APACyear {1947}}%
}]{%
mcnemar1947}
\APACinsertmetastar {%
mcnemar1947}%
\begin{APACrefauthors}%
McNemar, Q.%
\end{APACrefauthors}%
\unskip\
\newblock
\APACrefYearMonthDay{1947}{}{}.
\newblock
{\BBOQ}\APACrefatitle {Note on the Sampling Error of the Difference Between
  Correlated Proportions or Percentages} {Note on the sampling error of the
  difference between correlated proportions or percentages}.{\BBCQ}
\newblock
\APACjournalVolNumPages{Psychometrika}{12}{2}{153--157}.
\newblock
\begin{APACrefDOI} \doi{10.1007/BF02295996} \end{APACrefDOI}
\PrintBackRefs{\CurrentBib}

\bibitem [\protect \citeauthoryear {%
Mostafa%
, M{\"u}ller%
\BCBL {}\ \BBA {} Indiveri%
}{%
Mostafa%
\ \protect \BOthers {.}}{%
{\protect \APACyear {2015}}%
}]{%
mostafa2015}
\APACinsertmetastar {%
mostafa2015}%
\begin{APACrefauthors}%
Mostafa, H.%
, M{\"u}ller, L\BPBI K.%
\BCBL {}\ \BBA {} Indiveri, G.%
\end{APACrefauthors}%
\unskip\
\newblock
\APACrefYearMonthDay{2015}{}{}.
\newblock
{\BBOQ}\APACrefatitle {An Event-Based Architecture for Solving Constraint
  Satisfaction Problems} {An event-based architecture for solving constraint
  satisfaction problems}.{\BBCQ}
\newblock
\APACjournalVolNumPages{Nature Communications}{6}{}{8941}.
\newblock
\begin{APACrefDOI} \doi{10.1038/ncomms9941} \end{APACrefDOI}
\PrintBackRefs{\CurrentBib}

\bibitem [\protect \citeauthoryear {%
Neftci%
, Mostafa%
\BCBL {}\ \BBA {} Zenke%
}{%
Neftci%
\ \protect \BOthers {.}}{%
{\protect \APACyear {2019}}%
}]{%
neftci2019}
\APACinsertmetastar {%
neftci2019}%
\begin{APACrefauthors}%
Neftci, E\BPBI O.%
, Mostafa, H.%
\BCBL {}\ \BBA {} Zenke, F.%
\end{APACrefauthors}%
\unskip\
\newblock
\APACrefYearMonthDay{2019}{}{}.
\newblock
{\BBOQ}\APACrefatitle {Surrogate Gradient Learning in Spiking Neural Networks:
  Bringing the Power of Gradient-Based Optimization to Spiking Neural Networks}
  {Surrogate gradient learning in spiking neural networks: Bringing the power
  of gradient-based optimization to spiking neural networks}.{\BBCQ}
\newblock
\APACjournalVolNumPages{IEEE Signal Processing Magazine}{36}{6}{51--63}.
\newblock
\begin{APACrefDOI} \doi{10.1109/MSP.2019.2931595} \end{APACrefDOI}
\PrintBackRefs{\CurrentBib}

\bibitem [\protect \citeauthoryear {%
Palm%
, Paquet%
\BCBL {}\ \BBA {} Winther%
}{%
Palm%
\ \protect \BOthers {.}}{%
{\protect \APACyear {2018}}%
}]{%
palm2018}
\APACinsertmetastar {%
palm2018}%
\begin{APACrefauthors}%
Palm, R\BPBI B.%
, Paquet, U.%
\BCBL {}\ \BBA {} Winther, O.%
\end{APACrefauthors}%
\unskip\
\newblock
\APACrefYearMonthDay{2018}{}{}.
\newblock
{\BBOQ}\APACrefatitle {Recurrent Relational Networks} {Recurrent relational
  networks}.{\BBCQ}
\newblock
\BIn{} \APACrefbtitle {Advances in Neural Information Processing Systems}
  {Advances in neural information processing systems}\ (\BVOL~31).
\PrintBackRefs{\CurrentBib}

\bibitem [\protect \citeauthoryear {%
Paszke%
\ \protect \BOthers {.}}{%
Paszke%
\ \protect \BOthers {.}}{%
{\protect \APACyear {2019}}%
}]{%
paszke2019}
\APACinsertmetastar {%
paszke2019}%
\begin{APACrefauthors}%
Paszke, A.%
, Gross, S.%
, Massa, F.%
, Lerer, A.%
, Bradbury, J.%
, Chanan, G.%
\BDBL {}Chintala, S.%
\end{APACrefauthors}%
\unskip\
\newblock
\APACrefYearMonthDay{2019}{}{}.
\newblock
{\BBOQ}\APACrefatitle {{PyTorch}: An Imperative Style, High-Performance Deep
  Learning Library} {{PyTorch}: An imperative style, high-performance deep
  learning library}.{\BBCQ}
\newblock
\BIn{} \APACrefbtitle {Advances in Neural Information Processing Systems}
  {Advances in neural information processing systems}\ (\BVOL~32).
\PrintBackRefs{\CurrentBib}

\bibitem [\protect \citeauthoryear {%
Schwarzschild%
\ \protect \BOthers {.}}{%
Schwarzschild%
\ \protect \BOthers {.}}{%
{\protect \APACyear {2021}}%
}]{%
schwarzschild2021}
\APACinsertmetastar {%
schwarzschild2021}%
\begin{APACrefauthors}%
Schwarzschild, A.%
, Borgnia, E.%
, Gupta, A.%
, Huang, F.%
, Vishkin, U.%
, Goldblum, M.%
\BCBL {}\ \BBA {} Goldstein, T.%
\end{APACrefauthors}%
\unskip\
\newblock
\APACrefYearMonthDay{2021}{}{}.
\newblock
{\BBOQ}\APACrefatitle {Can You Learn an Algorithm? Generalizing from Easy to
  Hard Problems with Recurrent Networks} {Can you learn an algorithm?
  generalizing from easy to hard problems with recurrent networks}.{\BBCQ}
\newblock
\BIn{} \APACrefbtitle {Advances in Neural Information Processing Systems}
  {Advances in neural information processing systems}\ (\BVOL~34, \BPGS\
  6695--6706).
\PrintBackRefs{\CurrentBib}

\bibitem [\protect \citeauthoryear {%
Shah%
, Dikkala%
, Wang%
\BCBL {}\ \BBA {} Panigrahy%
}{%
Shah%
\ \protect \BOthers {.}}{%
{\protect \APACyear {2024}}%
}]{%
shah2024}
\APACinsertmetastar {%
shah2024}%
\begin{APACrefauthors}%
Shah, K.%
, Dikkala, N.%
, Wang, X.%
\BCBL {}\ \BBA {} Panigrahy, R.%
\end{APACrefauthors}%
\unskip\
\newblock
\APACrefYearMonthDay{2024}{}{}.
\newblock
{\BBOQ}\APACrefatitle {Causal Language Modeling Can Elicit Search and Reasoning
  Capabilities on Logic Puzzles} {Causal language modeling can elicit search
  and reasoning capabilities on logic puzzles}.{\BBCQ}
\newblock
\BIn{} \APACrefbtitle {Advances in Neural Information Processing Systems}
  {Advances in neural information processing systems}\ (\BVOL~37, \BPGS\
  56674--56702).
\PrintBackRefs{\CurrentBib}

\bibitem [\protect \citeauthoryear {%
Stimberg%
, Brette%
\BCBL {}\ \BBA {} Goodman%
}{%
Stimberg%
\ \protect \BOthers {.}}{%
{\protect \APACyear {2019}}%
}]{%
stimberg2019}
\APACinsertmetastar {%
stimberg2019}%
\begin{APACrefauthors}%
Stimberg, M.%
, Brette, R.%
\BCBL {}\ \BBA {} Goodman, D\BPBI F\BPBI M.%
\end{APACrefauthors}%
\unskip\
\newblock
\APACrefYearMonthDay{2019}{}{}.
\newblock
{\BBOQ}\APACrefatitle {{Brian} 2, an Intuitive and Efficient Neural Simulator}
  {{Brian} 2, an intuitive and efficient neural simulator}.{\BBCQ}
\newblock
\APACjournalVolNumPages{eLife}{8}{}{e47314}.
\newblock
\begin{APACrefDOI} \doi{10.7554/eLife.47314} \end{APACrefDOI}
\PrintBackRefs{\CurrentBib}

\bibitem [\protect \citeauthoryear {%
Tang%
, Kumar%
, Yoo%
\BCBL {}\ \BBA {} Michmizos%
}{%
Tang%
\ \protect \BOthers {.}}{%
{\protect \APACyear {2021}}%
}]{%
tang2021popsan}
\APACinsertmetastar {%
tang2021popsan}%
\begin{APACrefauthors}%
Tang, G.%
, Kumar, N.%
, Yoo, R.%
\BCBL {}\ \BBA {} Michmizos, K\BPBI P.%
\end{APACrefauthors}%
\unskip\
\newblock
\APACrefYearMonthDay{2021}{}{}.
\newblock
{\BBOQ}\APACrefatitle {Deep Reinforcement Learning with Population-Coded
  Spiking Neural Network for Continuous Control} {Deep reinforcement learning
  with population-coded spiking neural network for continuous control}.{\BBCQ}
\newblock
\BIn{} \APACrefbtitle {Proceedings of the 2020 Conference on Robot Learning}
  {Proceedings of the 2020 conference on robot learning}\ (\BVOL~155, \BPGS\
  2016--2029).
\PrintBackRefs{\CurrentBib}

\bibitem [\protect \citeauthoryear {%
Veli{\v{c}}kovi{\'c}%
\ \protect \BOthers {.}}{%
Veli{\v{c}}kovi{\'c}%
\ \protect \BOthers {.}}{%
{\protect \APACyear {2022}}%
}]{%
velickovic2022}
\APACinsertmetastar {%
velickovic2022}%
\begin{APACrefauthors}%
Veli{\v{c}}kovi{\'c}, P.%
, Badia, A\BPBI P.%
, Budden, D.%
, Pascanu, R.%
, Banino, A.%
, Dashevskiy, M.%
\BDBL {}Blundell, C.%
\end{APACrefauthors}%
\unskip\
\newblock
\APACrefYearMonthDay{2022}{}{}.
\newblock
{\BBOQ}\APACrefatitle {The {CLRS} Algorithmic Reasoning Benchmark} {The {CLRS}
  algorithmic reasoning benchmark}.{\BBCQ}
\newblock
\BIn{} \APACrefbtitle {Proceedings of the 39th International Conference on
  Machine Learning} {Proceedings of the 39th international conference on
  machine learning}\ (\BVOL~162, \BPGS\ 22084--22102).
\PrintBackRefs{\CurrentBib}

\bibitem [\protect \citeauthoryear {%
G.~Wang%
\ \protect \BOthers {.}}{%
G.~Wang%
\ \protect \BOthers {.}}{%
{\protect \APACyear {2025}}%
}]{%
wang2025hrm}
\APACinsertmetastar {%
wang2025hrm}%
\begin{APACrefauthors}%
Wang, G.%
, Li, J.%
, Sun, Y.%
, Chen, X.%
, Liu, C.%
, Wu, Y.%
\BDBL {}{Abbasi Yadkori}, Y.%
\end{APACrefauthors}%
\unskip\
\newblock
\APACrefYearMonthDay{2025}{}{}.
\newblock
{\BBOQ}\APACrefatitle {Hierarchical Reasoning Model} {Hierarchical reasoning
  model}.{\BBCQ}
\newblock
\APACjournalVolNumPages{arXiv preprint arXiv:2506.21734}{}{}{}.
\newblock
\begin{APACrefDOI} \doi{10.48550/arXiv.2506.21734} \end{APACrefDOI}
\PrintBackRefs{\CurrentBib}

\bibitem [\protect \citeauthoryear {%
P\BHBI W.~Wang%
, Donti%
, Wilder%
\BCBL {}\ \BBA {} Kolter%
}{%
P\BHBI W.~Wang%
\ \protect \BOthers {.}}{%
{\protect \APACyear {2019}}%
}]{%
wang2019}
\APACinsertmetastar {%
wang2019}%
\begin{APACrefauthors}%
Wang, P\BHBI W.%
, Donti, P\BPBI L.%
, Wilder, B.%
\BCBL {}\ \BBA {} Kolter, J\BPBI Z.%
\end{APACrefauthors}%
\unskip\
\newblock
\APACrefYearMonthDay{2019}{}{}.
\newblock
{\BBOQ}\APACrefatitle {{SATNet}: Bridging Deep Learning and Logical Reasoning
  Using a Differentiable Satisfiability Solver} {{SATNet}: Bridging deep
  learning and logical reasoning using a differentiable satisfiability
  solver}.{\BBCQ}
\newblock
\BIn{} \APACrefbtitle {Proceedings of the 36th International Conference on
  Machine Learning} {Proceedings of the 36th international conference on
  machine learning}\ (\BVOL~97, \BPGS\ 6545--6554).
\PrintBackRefs{\CurrentBib}

\bibitem [\protect \citeauthoryear {%
Wies%
, Levine%
\BCBL {}\ \BBA {} Shashua%
}{%
Wies%
\ \protect \BOthers {.}}{%
{\protect \APACyear {2023}}%
}]{%
wies2023}
\APACinsertmetastar {%
wies2023}%
\begin{APACrefauthors}%
Wies, N.%
, Levine, Y.%
\BCBL {}\ \BBA {} Shashua, A.%
\end{APACrefauthors}%
\unskip\
\newblock
\APACrefYearMonthDay{2023}{}{}.
\newblock
{\BBOQ}\APACrefatitle {Sub-Task Decomposition Enables Learning in Sequence to
  Sequence Tasks} {Sub-task decomposition enables learning in sequence to
  sequence tasks}.{\BBCQ}
\newblock
\BIn{} \APACrefbtitle {Proceedings of the 11th International Conference on
  Learning Representations ({ICLR}).} {Proceedings of the 11th international
  conference on learning representations ({ICLR}).}
\PrintBackRefs{\CurrentBib}

\bibitem [\protect \citeauthoryear {%
Williams%
}{%
Williams%
}{%
{\protect \APACyear {1992}}%
}]{%
williams1992}
\APACinsertmetastar {%
williams1992}%
\begin{APACrefauthors}%
Williams, R\BPBI J.%
\end{APACrefauthors}%
\unskip\
\newblock
\APACrefYearMonthDay{1992}{}{}.
\newblock
{\BBOQ}\APACrefatitle {Simple Statistical Gradient-Following Algorithms for
  Connectionist Reinforcement Learning} {Simple statistical gradient-following
  algorithms for connectionist reinforcement learning}.{\BBCQ}
\newblock
\APACjournalVolNumPages{Machine Learning}{8}{3--4}{229--256}.
\newblock
\begin{APACrefDOI} \doi{10.1007/BF00992696} \end{APACrefDOI}
\PrintBackRefs{\CurrentBib}

\bibitem [\protect \citeauthoryear {%
Xu%
\ \protect \BOthers {.}}{%
Xu%
\ \protect \BOthers {.}}{%
{\protect \APACyear {2020}}%
}]{%
xu2020}
\APACinsertmetastar {%
xu2020}%
\begin{APACrefauthors}%
Xu, K.%
, Li, J.%
, Zhang, M.%
, Du, S\BPBI S.%
, Kawarabayashi, K.%
\BCBL {}\ \BBA {} Jegelka, S.%
\end{APACrefauthors}%
\unskip\
\newblock
\APACrefYearMonthDay{2020}{}{}.
\newblock
{\BBOQ}\APACrefatitle {What Can Neural Networks Reason About?} {What can neural
  networks reason about?}{\BBCQ}
\newblock
\BIn{} \APACrefbtitle {Proceedings of the 8th International Conference on
  Learning Representations ({ICLR}).} {Proceedings of the 8th international
  conference on learning representations ({ICLR}).}
\PrintBackRefs{\CurrentBib}

\bibitem [\protect \citeauthoryear {%
Yang%
, Ishay%
\BCBL {}\ \BBA {} Lee%
}{%
Yang%
\ \protect \BOthers {.}}{%
{\protect \APACyear {2023}}%
}]{%
yang2023}
\APACinsertmetastar {%
yang2023}%
\begin{APACrefauthors}%
Yang, Z.%
, Ishay, A.%
\BCBL {}\ \BBA {} Lee, J.%
\end{APACrefauthors}%
\unskip\
\newblock
\APACrefYearMonthDay{2023}{}{}.
\newblock
{\BBOQ}\APACrefatitle {Learning to Solve Constraint Satisfaction Problems with
  Recurrent Transformer} {Learning to solve constraint satisfaction problems
  with recurrent transformer}.{\BBCQ}
\newblock
\BIn{} \APACrefbtitle {Proceedings of the 11th International Conference on
  Learning Representations ({ICLR}).} {Proceedings of the 11th international
  conference on learning representations ({ICLR}).}
\PrintBackRefs{\CurrentBib}

\bibitem [\protect \citeauthoryear {%
Zenke%
\ \BBA {} Ganguli%
}{%
Zenke%
\ \BBA {} Ganguli%
}{%
{\protect \APACyear {2018}}%
}]{%
zenke2018}
\APACinsertmetastar {%
zenke2018}%
\begin{APACrefauthors}%
Zenke, F.%
\BCBT {}\ \BBA {} Ganguli, S.%
\end{APACrefauthors}%
\unskip\
\newblock
\APACrefYearMonthDay{2018}{}{}.
\newblock
{\BBOQ}\APACrefatitle {{SuperSpike}: Supervised Learning in Multilayer Spiking
  Neural Networks} {{SuperSpike}: Supervised learning in multilayer spiking
  neural networks}.{\BBCQ}
\newblock
\APACjournalVolNumPages{Neural Computation}{30}{6}{1514--1541}.
\newblock
\begin{APACrefDOI} \doi{10.1162/neco_a_01086} \end{APACrefDOI}
\PrintBackRefs{\CurrentBib}

\bibitem [\protect \citeauthoryear {%
Zenke%
\ \BBA {} Vogels%
}{%
Zenke%
\ \BBA {} Vogels%
}{%
{\protect \APACyear {2021}}%
}]{%
zenke2021}
\APACinsertmetastar {%
zenke2021}%
\begin{APACrefauthors}%
Zenke, F.%
\BCBT {}\ \BBA {} Vogels, T\BPBI P.%
\end{APACrefauthors}%
\unskip\
\newblock
\APACrefYearMonthDay{2021}{}{}.
\newblock
{\BBOQ}\APACrefatitle {The Remarkable Robustness of Surrogate Gradient Learning
  for Instilling Complex Function in Spiking Neural Networks} {The remarkable
  robustness of surrogate gradient learning for instilling complex function in
  spiking neural networks}.{\BBCQ}
\newblock
\APACjournalVolNumPages{Neural Computation}{33}{4}{899--925}.
\newblock
\begin{APACrefDOI} \doi{10.1162/neco_a_01367} \end{APACrefDOI}
\PrintBackRefs{\CurrentBib}

\end{thebibliography}

\end{document}